\documentclass[journal,10pt,twoside]{IEEEtran}

\usepackage{cite}
\usepackage{amsmath,amssymb}
\usepackage{graphicx}
\usepackage{placeins}
\usepackage{booktabs}
\usepackage{array}
\usepackage{multirow}
\usepackage{xcolor}
\usepackage{url}
\usepackage{hyperref}
\usepackage{balance}
\usepackage{enumitem}
\usepackage{pifont}    
\usepackage{siunitx}   
\usepackage{tikz}
\usetikzlibrary{arrows.meta,calc,positioning,shapes.geometric,
                shapes.multipart,fit,backgrounds,
                decorations.pathreplacing,decorations.markings}
\usepackage{pgfplots}
\pgfplotsset{compat=1.18}
\usepgfplotslibrary{groupplots,statistics}

\definecolor{himecInk}{HTML}{1D2730}
\definecolor{himecMuted}{HTML}{66707A}
\definecolor{himecGrid}{HTML}{CBD2D8}
\definecolor{himecRed}{HTML}{EA6B66}
\definecolor{himecBlue}{HTML}{3399FF}
\definecolor{himecGreen}{HTML}{91935D}
\definecolor{himecPurple}{HTML}{8E67C7}

\pgfplotsset{
  himec/.style={
    width=0.98\linewidth, height=5.8cm,
    ymin=0,
    grid=major, grid style={draw=himecGrid,thin},
    axis line style={draw=himecInk},
    tick align=outside,
    tick label style={font=\scriptsize,text=himecInk},
    label style={font=\small,text=himecInk},
    legend style={font=\scriptsize, draw=himecGrid, fill=white,
                  inner sep=2pt, row sep=-1pt},
    every axis plot/.append style={line width=0.7pt},
  }
}

\graphicspath{
  {./}
  {outputs/mismatch_analysis/}
  {outputs/aadd_visualization/}
  {outputs/qualitative/}
  {outputs/qualitative/compare/}
  {outputs/v4_ecsd/analysis/}
  {../HIMEC_FZ_NCC_Combined/figures/himec/}
  {../IEEE_TGRS_Submission/HIMEC/figures/}
  {../IEEE_TGRS_Submission/HIMEC/results/}
}

\hypersetup{hidelinks}
\newcommand{\tstrut}{\renewcommand{\arraystretch}{1.15}}

\definecolor{capok}{RGB}{0,140,60}
\definecolor{caperr}{RGB}{200,30,30}
\newcommand{\cok}[1]{\textbf{\textcolor{capok}{#1}}}
\newcommand{\cerr}[1]{\textit{\textcolor{caperr}{#1}}}

\begin{document}

\title{HIMEC: Directional Change Representation and Fixed-Interface Decoding for Remote Sensing Image Change Captioning}

\author{Aysha~Ashraf,~Shaina~Ashraf,~Wafaa~I.~M.~Hussin,~Ali~Haider,~Zhi~Lu,~and~Zhenming~Peng,~\IEEEmembership{Senior~Member,~IEEE}%
\thanks{This work was supported by the Natural Science Foundation of Sichuan
Province of China (Grant No.\ 2025ZNSFSC0522) and the National Natural
Science Foundation of China (Grant No.\ 61571096).
(\emph{Corresponding authors: Zhi Lu and Zhenming Peng.})}%
\thanks{A.~Ashraf, W.~I.~M.~Hussin, A.~Haider, and Z.~Peng are with the School
of Information and Communication Engineering, Laboratory of Imaging Detection and
Intelligent Perception, University of Electronic Science and Technology of China,
Chengdu 611731, China
(e-mail: aysha.ashraf92@gmail.com; wafaaibrahim20@gmail.com;
alihaider681@hotmail.com; zmpeng@uestc.edu.cn).}%
\thanks{S.~Ashraf is with the Bonn-Aachen International Center for Information
Technology (B-IT), University of Bonn, Bonn, Germany
(e-mail: shainamscs@gmail.com).}%
\thanks{Z.~Lu is with the Laboratory of Intelligent Collaborative Computing,
University of Electronic Science and Technology of China, Chengdu 611731,
China (e-mail: zhilu@uestc.edu.cn).}}

\markboth{IEEE TRANSACTIONS ON GEOSCIENCE AND REMOTE SENSING}%
{}

\maketitle
\begin{abstract}
Remote sensing image change captioning (RSICC) converts bitemporal imagery into a sentence describing semantic changes. Most RSICC methods condition caption decoders directly on fused visual features, leaving intermediate change structure and decoder-interface consistency less studied. We present HIMEC, a framework that combines Directional Change Representation (DCR) with fixed-interface decoding. DCR separates signed differences into appearance-oriented, disappearance-oriented, and shared-context streams before fusion. A learned-query encoder converts the fused representation into visually conditioned change-query tokens that form the scene decoder's only sample-dependent memory. A training-only auxiliary phrase decoder supplies caption-derived supervision. With a fixed zero input, the scene decoder maintains the same interface during training and inference. Separately, we evaluate a local-to-scene cascade conditioned on teacher-forced local states during training and autoregressive states at inference. On changed LEVIR-CC validation pairs, these states have a mean cosine distance of 0.69. Regime-matched conditioning recovers most of the associated deficit, whereas permuting state correspondence causes no detectable penalty. These findings are limited to the evaluated cascade. In a matched three-seed comparison, HIMEC reaches a Consensus-based Image Description Evaluation (CIDEr) score of $142.81\pm0.60$ on LEVIR-CC, versus $139.51\pm3.40$ for direct fused-feature memory. On SECOND-CC, fixed-zero and regime-matched diagnostic conditioning reach 75.67 and 76.99 CIDEr, respectively, versus 60.77 for the mismatched cascade. The source code will be made publicly available at \url{https://github.com/ayshaashra/HIMEC} upon publication.
\end{abstract}


\section{Introduction}

\IEEEPARstart{R}{emote} sensing change analysis compares co-registered images acquired at different times to support urban monitoring, disaster assessment, agricultural management, and environmental surveillance. Most methods produce binary or semantic maps that localize changed regions. Deep learning has advanced these methods from convolutional Siamese networks to Transformer architectures~\cite{chen2021bit,bandara2022changeformer} and difference-aware or edge-guided refinement strategies~\cite{li2024stadecdnet,hussin2026edgerefnet}. Although change maps localize and categorize changes, they do not express scene-level change semantics in open-vocabulary language.

Remote sensing image change captioning (RSICC) addresses this limitation by generating a natural-language description from a bitemporal image pair~\cite{rsicsurvey2025}. LEVIR-CC established the first large-scale RSICC benchmark~\cite{liu2022rsicformer}, and subsequent work has improved change encoding, language decoding, auxiliary semantic guidance, and computational efficiency~\cite{chang2023chg2cap,liu2024multitask,rscama2024}. Despite these variations, most methods encode the two images, form a visual representation of their differences, and condition a single autoregressive decoder on that representation.

Prior work uses change-specific attention~\cite{liu2022mccformer}, pixel-level supervision, and semantic auxiliary guidance~\cite{liu2024multitask,semanticcc2024} to structure visual evidence. However, most models still pass a single fused visual memory directly to the caption decoder. This design makes temporal direction implicit and leaves the organization of distinct change cues to the decoder. We therefore examine whether directional features and a compact learned-query interface can structure change evidence before sentence generation. The interface is evaluated as a whole; individual tokens are not interpreted as annotated objects or regions.

We propose HIMEC, which combines Directional Change Representation (DCR) with a learned-query interface. DCR separates appearance-oriented, disappearance-oriented, and shared-context evidence before fusion. Learned queries organize the fused evidence as visually conditioned change-query tokens. A training-only auxiliary decoder predicts caption-derived change phrases, while the scene decoder generates the caption through a fixed zero interface. Auxiliary-decoder states are not transferred to the scene decoder.

Separately, we analyze a local-to-scene cascade whose downstream decoder receives teacher-forced local states during training and autoregressive states at inference. Regime-matched and content-permuted controls test whether its performance is associated with conditioning inconsistency or image-specific local-state content. This diagnostic cascade is not part of the HIMEC inference path.

The main contributions are summarized as follows.

\begin{enumerate}
\item We introduce and directly test a structured change-query interface for RSICC. With matched DCR and training, change-query memory obtains $142.81\pm0.60$ CIDEr versus $139.51\pm3.40$ for direct fused-feature memory, without transferring autoregressive local-decoder states at inference.

\item We organize signed temporal evidence with DCR and use caption-derived change phrases only as training-time auxiliary targets. Configuration-matched ablations show score reductions after removing the difference interaction, appearance stream, or shared-context stream; the other tested removals do not lower the observed score.

\item In a separate diagnostic study, we quantify the training-to-inference discrepancy in the evaluated local-to-scene cascades and test regime-matched and content-permuted controls. We evaluate these controls on LEVIR-CC, SECOND-CC, DUBAI-CC, an independent encoder-decoder implementation, and a 300-pair LEVIR-CC subset.
\end{enumerate}

The remainder of this paper is organized as follows. Section~\ref{sec:related} reviews related work, Section~\ref{sec:method} presents the proposed method, Section~\ref{sec:setup} describes the experiments, and Section~\ref{sec:conclusion} concludes the paper.

\section{Related Work}
\label{sec:related}

\subsection{Remote Sensing Change Detection}

Recent methods improve localization through spatiotemporal attention~\cite{li2024stadecdnet}, edge-guided refinement~\cite{hussin2026edgerefnet,glfer2025}, uncertainty-aware learning~\cite{uncertaintycd2025}, and optimization strategies~\cite{bethechange2024}. These map-producing approaches are complementary to change captioning, which describes scene changes in open-vocabulary language.
\subsection{Common RSICC Paradigm}
The common RSICC pipeline contains three stages. First, a shared or bilateral encoder extracts ordered feature maps from the pre-change and post-change images. Second, a change-representation module combines these maps through concatenation, differencing, attention, or semantic guidance to form a visual memory. Third, an autoregressive language decoder conditions on that memory and predicts a scene-level caption. In compact form, the prevailing design is
\begin{equation}
(I_1,I_2)\xrightarrow{\mathcal{E}}(F_1,F_2)
\xrightarrow{\mathcal{C}}M_{\mathrm{chg}}
\xrightarrow{\mathcal{D}}Y,
\label{eq:common_rsicc}
\end{equation}
where $\mathcal{E}$, $\mathcal{C}$, and $\mathcal{D}$ denote image encoding, change representation, and caption decoding, respectively. Early methods used globally pooled convolutional features with recurrent decoders~\cite{cai2021duda,liu2023changes,hoxha2022change}. RSICCformer combines bitemporal and difference features with Transformer encoding and decoding~\cite{liu2022rsicformer}; Chg2Cap uses attentive change features with a Transformer caption decoder~\cite{chang2023chg2cap}; SAT-Cap refines symmetric differences through spatial attention~\cite{yao2025satcap}; and RSCaMa replaces conventional sequence modeling with state-space blocks~\cite{rscama2024}.

Subsequent studies enrich one or more stages of this pipeline. MCCFormer models and localizes multiple changes through change-specific attention~\cite{liu2022mccformer}; other methods add joint detection-captioning objectives~\cite{liu2024multitask}, semantic or pixel-level guidance~\cite{semanticcc2024,karaca2025mmodalcc,samguided2025}, concept interaction~\cite{icnt2025}, semantic-spatial modeling~\cite{sscp2025}, retrieval augmentation~\cite{ragcc2025}, or instruction tuning~\cite{keychange2025}. Large vision-language models provide another route~\cite{ccexpert2024,deltavlm2025}, while lightweight and state-space variants target computational cost~\cite{lightweightdct2025,ctsdnet2025,pm3net2026,mthnet2025}.

These methods share the direct visual-memory-to-caption design in~\eqref{eq:common_rsicc}.
\subsection{Learned Visual Queries and Operational Terminology}
Learned queries provide a fixed set of latent slots that can attend to visual memory, as demonstrated by DETR~\cite{carion2020detr}. Their meaning nevertheless depends on the supervision and evaluation used in a given task. Visual grounding conventionally requires an explicit or evaluated correspondence between language and image regions. For example, MDETR uses phrase--object alignment supervision and evaluates grounding tasks~\cite{kamath2021mdetr}, whereas Align2Ground learns latent phrase--region correspondences and evaluates phrase localization~\cite{datta2019align2ground}.
\subsection{Bitemporal Difference Representation}
RSICC methods construct bitemporal features through ordered concatenation, absolute differences, or attention-enhanced change maps. RSICCformer~\cite{liu2022rsicformer} combines pre-change, post-change, and absolute-difference features. Chg2Cap~\cite{chang2023chg2cap} integrates change-attention maps into decoding, whereas SAT-Cap~\cite{yao2025satcap} refines symmetric difference features with spatial attention.
\subsection{Train--Inference Consistency in Cascaded Decoders}
Exposure bias arises when a decoder is trained with ground-truth histories but conditions on its own predictions at inference. Common remedies include scheduled sampling~\cite{bengio2015scheduled}, sequence-level reinforcement learning~\cite{ranzato2016seqlevel}, and professor forcing~\cite{lamb2016professor}. A cascaded dual-decoder architecture introduces an additional interface discrepancy when the downstream decoder receives teacher-forced local states during training and autoregressive states during inference. This problem is related to the covariate shift studied in cascaded prediction and imitation learning~\cite{ross2011dagger}. The RSICC studies reviewed above do not report paired local-state measurements or controls that vary both the conditioning regime and image-state correspondence.
\subsection{Self-Critical Sequence Training}
Self-Critical Sequence Training (SCST)~\cite{rennie2017scst} applies a REINFORCE policy gradient with a greedy-decoding baseline to optimize sequence-level metrics such as CIDEr~\cite{vedantam2015cider}. SCST has been used as a second training stage in RSICC~\cite{liu2022rsicformer,chang2023chg2cap,yao2025satcap}.

HIMEC retains the encoder-representation-decoder organization in~\eqref{eq:common_rsicc} but inserts a fixed learned-query interface between $M_{\mathrm{chg}}$ and the scene decoder. Its DCR module preserves the direction of signed feature differences together with shared context. HIMEC does not use bounding boxes, masks, region proposals, object identities, or phrase--region annotations. We therefore refer to its latent representations as \emph{visually conditioned change-query tokens}. The resulting cross-attention weights are visualized only as diagnostic attention maps and are not evaluated as spatial or semantic alignments. HIMEC applies SCST after cross-entropy training as a caption-level optimization stage; SCST does not address the inter-decoder conditioning discrepancy. The local-to-scene cascade analyzed later is a separate diagnostic architecture and is not part of the HIMEC inference path.

\section{Method}
\label{sec:method}

\subsection{Problem Formulation and Framework Overview}
\label{sec:overview}
\label{sec:problem}
\begin{figure*}[!t]
\centering
\includegraphics[width=\textwidth,keepaspectratio]{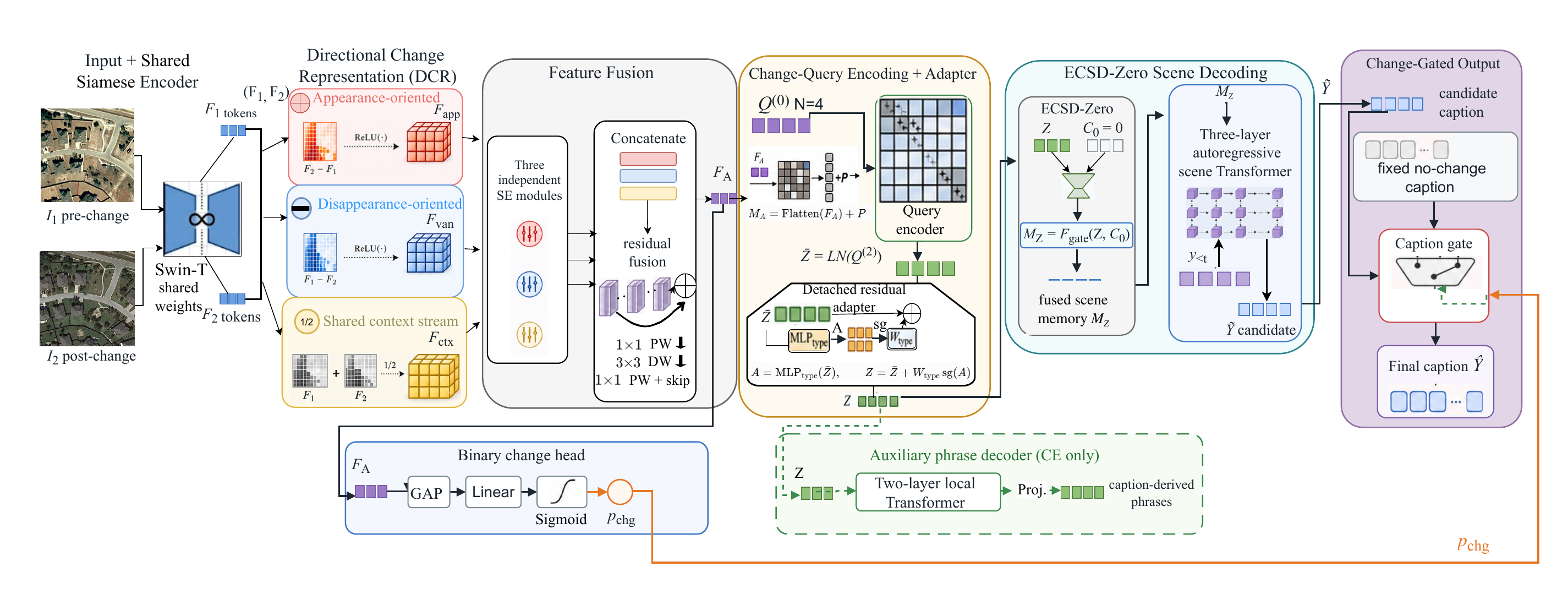}
\caption{Architecture of HIMEC. The shared Siamese encoder, DCR, and learned-query encoder transform the image pair into change-query tokens $Z$. The scene decoder uses $M_Z=\mathcal F_{\mathrm{gate}}(Z,C_0)$, with $C_0=\mathbf 0$, during CE training, SCST, and inference. The dashed green branch is a CE-only auxiliary phrase decoder; during SCST, only the scene-generation path is active. The binary change head is optimized during CE and controls the caption gate only at inference. Local-decoder hidden states are never supplied to the scene decoder.}
\label{fig:architecture}
\end{figure*}

Given two co-registered remote sensing images $I_1$ and $I_2$ of the same area acquired at different times, the task is to generate a caption $Y=(y_1,\ldots,y_L)$ that describes their semantic changes. A \emph{caption-derived change phrase} is a parser-produced fragment containing a change-related verb, or the complete reference caption used as a fallback when no such fragment is found. It is a linguistic auxiliary target and supplies neither object identity nor a spatial boundary. A \emph{change-query token} is a latent vector produced when a learned query attends to fused bitemporal features. It is not assumed to correspond one-to-one with an annotated object, region, or discrete semantic change.
HIMEC maps the image pair to a matrix of $N$ visually conditioned change-query tokens and composes a scene-level caption from their joint representation:
\begin{align}
Z=[z_1,\ldots,z_N]^{\top}
&=\phi_{\theta}(I_1,I_2)
\in\mathbb{R}^{N\times d},
\label{eq:queries}\\
\widetilde Y&=g_{\psi}(Z).
\label{eq:compose}
\end{align}
where $\widetilde Y$ denotes the candidate scene caption before the final change/no-change decision. A shared-weight Siamese Swin Transformer Tiny (Swin-T) encoder followed by a learned feature projection produces $F_1,F_2\in\mathbb{R}^{d\times16\times16}$, with the batch dimension omitted. Directional Change Representation (DCR) fuses these maps into $F_A$, and learned-query cross-attention summarizes $F_A$ as $Z$. During cross-entropy training, $Z$ supports an auxiliary caption-derived phrase objective and the primary scene-caption objective. Fig.~\ref{fig:architecture} summarizes the resulting data flow and component schedule.

\subsection{Change Representation and Query-Based Change Encoding}
\label{sec:change_rep}

\subsubsection{Directional Change Representation}
\label{sec:aadd}

DCR separates the two directions of the signed feature difference before fusion. Given $F_1$ and $F_2$, it retains the positive part of each direction together with the temporal mean:
\begin{align}
F_{\mathrm{App}}
&=\operatorname{ReLU}(F_2-F_1),
\label{eq:appear}\\
F_{\mathrm{Van}}
&=\operatorname{ReLU}(F_1-F_2),
\label{eq:vanish}\\
F_{\mathrm{Ctx}}
&=\frac{F_1+F_2}{2}.
\label{eq:context}
\end{align}
$F_{\mathrm{App}}$ and $F_{\mathrm{Van}}$ retain the two positive directions of the feature difference, whereas $F_{\mathrm{Ctx}}$ is a symmetric temporal average rather than a supervised unchanged-region map. Before learned reweighting and fusion, the three streams form a deterministic reparameterization of $(F_1,F_2)$.

Each stream is processed by an independent squeeze-and-excitation (SE) block~\cite{hu2018se}. The reweighted maps are concatenated and fused by a depthwise-separable residual bottleneck~\cite{chollet2017xception}:
\begin{align}
\widetilde F_s
&=\operatorname{SE}_s(F_s),
\quad s\in\{\mathrm{App},\mathrm{Van},\mathrm{Ctx}\},
\label{eq:dcr_se}\\
X
&=\mathcal P_{\mathrm{in}}
\left(
[\widetilde F_{\mathrm{App}};
 \widetilde F_{\mathrm{Van}};
 \widetilde F_{\mathrm{Ctx}}]
\right),
\label{eq:dcr_concat}\\
F_A
&=\operatorname{ReLU}
\left(
\mathcal P_{\mathrm{out}}(\mathcal D(X))+X
\right).
\label{eq:dcr_fusion}
\end{align}
Here $[\,;\,]$ denotes channel concatenation. $\mathcal P_{\mathrm{in}}$ is a $1{\times}1$ convolution followed by batch normalization and ReLU; $\mathcal D$ is a $3{\times}3$ depthwise convolution followed by batch normalization and ReLU; and $\mathcal P_{\mathrm{out}}$ is a $1{\times}1$ convolution followed by batch normalization. Only the signed-direction decomposition in \eqref{eq:appear}--\eqref{eq:context} is parameter-free; the SE blocks and residual fusion are learned. The output is $F_A\in\mathbb{R}^{d\times16\times16}$ with $d=512$.

A separate change head applies global average pooling (GAP) and a linear layer to $F_A$:
\begin{equation}
s_{\mathrm{chg}}
=w_c^{\top}\operatorname{GAP}(F_A)+b_c,
\label{eq:change_logit}
\end{equation}
where $s_{\mathrm{chg}}$ is the change logit. This logit is used directly by the binary change-classification loss. At inference, $p_{\mathrm{chg}}=\sigma(s_{\mathrm{chg}})$ is compared with the applicable threshold specified in Section~\ref{sec:metrics}.

\subsubsection{Change-Query Encoding}
\label{sec:query_encoding}

Let $Q^{(0)}=[q_1,\ldots,q_N]^{\top}\in\mathbb{R}^{N\times d}$ denote the learned query embeddings. The visual memory is
\begin{equation}
M_A=\operatorname{Flatten}_{hw}(F_A)+P
\in\mathbb{R}^{256\times d},
\label{eq:visual_memory}
\end{equation}
where $\operatorname{Flatten}_{hw}$ converts the $16{\times}16$ grid into a channel-last token sequence and $P\in\mathbb{R}^{256\times d}$ is a learned positional embedding. The reported model uses $N=4$, two Transformer query-encoding layers, and eight attention heads:
\begin{equation}
\begin{aligned}
Q^{(\ell)}
&=\operatorname{DecLayer}_{\ell}
\left(Q^{(\ell-1)},M_A\right),
\quad \ell\in\{1,2\},\\
\bar Z
&=\operatorname{LN}\left(Q^{(2)}\right).
\end{aligned}
\label{eq:query_encoding}
\end{equation}
Here $\operatorname{LN}$ denotes layer normalization. Each layer contains query self-attention, standard multihead cross-attention over $M_A$, and a feed-forward network. The fixed-cardinality learned query set provides a compact interface whose slots can attend to different portions of the visual memory while being optimized jointly for scene captioning. The head-averaged final-layer cross-attention weights yield one soft $16{\times}16$ response map per query. We visualize these weights only as diagnostic attention maps; without spatial supervision or localization evaluation, they are not interpreted as phrase--region alignments.

The reported checkpoints retain the following detached residual adapter:
\begin{equation}
\begin{aligned}
A&=\operatorname{MLP}_{\mathrm{type}}(\bar Z)
\in\mathbb{R}^{N\times K},\\
Z&=\bar Z+\operatorname{sg}(A)W_{\mathrm{type}}^{\top},
\quad W_{\mathrm{type}}\in\mathbb{R}^{d\times K}.
\end{aligned}
\label{eq:type_adapter}
\end{equation}
where $\operatorname{sg}(\cdot)$ stops gradients through $A$ and $K=5$ is the checkpoint classifier width. The semantic change-type loss is disabled in all reported experiments $(\lambda_{\mathrm{type}}=0)$; therefore, the $\operatorname{MLP}_{\mathrm{type}}$ parameters remain fixed at initialization, whereas $W_{\mathrm{type}}$ is optimized by the captioning objectives. We retain the adapter for checkpoint fidelity and do not treat it as a semantic predictor or claimed contribution. Its effect is not isolated by the current ablations. During cross-entropy training, the queries receive gradients from the scene, local, and diversity losses; during SCST, only the scene-generation path updates them.

\subsection{Auxiliary Phrase Supervision and Scene Decoding}
\label{sec:phrase_supervision}

\subsubsection{Auxiliary Change-Phrase Supervision}

The change-query tokens $Z$ provide the shared representation for the phrase and scene objectives during cross-entropy training. The local Transformer decoder predicts an auxiliary caption-derived change phrase $E_k=(e_{k,1},\ldots,e_{k,T_k})$ from each token $z_k$. Under teacher forcing, it models
\begin{equation}
p_{\mathrm{local}}(E_k\mid z_k)
=
\prod_{t=1}^{T_k}
p(e_{k,t}\mid e_{k,<t},z_k).
\label{eq:local_caption}
\end{equation}

The scene-caption target and its auxiliary phrase targets are derived from the same selected reference. After lowercasing, the rule-based parser splits the caption at ``and,'' ``while,'' commas, and semicolons. It retains fragments containing an entry in a fixed change-verb lexicon; if none remains, it uses the complete caption as a fallback target. The parser returns at most $N$ phrases in extraction order. Unfilled slots receive all-padding targets and are excluded from the local loss. During training, the first reference yielding the maximum number of extracted phrases is used with probability 0.8; otherwise, a reference is sampled uniformly. The parser maps unchanged samples to ``no change,'' but the implemented local loss includes only valid phrase slots from changed samples. The local objective provides slot-indexed linguistic supervision, but the configuration-matched \texttt{no\_local} comparison does not establish a captioning gain from retaining this auxiliary decoder.

\subsubsection{Scene Decoder With a Fixed Interface}
\label{sec:ecsd}

The scene decoder contains a two-input gated fusion block. For change-query tokens $Z$ and a second input $C$ of the same shape, the block computes
\begin{align}
G(Z,C)
&=\sigma\left([Z;C]W_g^{\top}
+\mathbf 1_Nb_g^{\top}\right),
\nonumber\\
\mathcal F_{\mathrm{gate}}(Z,C)
&=
\operatorname{LN}
\Big(
\operatorname{Drop}\big[
G(Z,C)\odot
\nonumber\\
&\hspace{12mm}
\tanh\left(ZW_e^{\top}+\mathbf 1_Nb_e^{\top}\right)
\nonumber\\
&\hspace{4mm}
+(\mathbf 1_{N\times d}-G(Z,C))\odot
\nonumber\\
&\hspace{12mm}
\tanh\left(CW_h^{\top}+\mathbf 1_Nb_h^{\top}\right)
\big]
\Big),
\label{eq:gated_fusion}
\end{align}
where $[Z;C]\in\mathbb R^{N\times2d}$ denotes feature-wise concatenation, $W_g\in\mathbb R^{d\times2d}$, $W_e,W_h\in\mathbb R^{d\times d}$, and $b_g,b_e,b_h\in\mathbb R^d$. All affine maps act row-wise; $\odot$ and $\operatorname{Drop}$ denote elementwise multiplication and dropout, respectively.

In HIMEC, the second input is fixed to zero during both training and inference:
\begin{equation}
C_0=\mathbf 0_{N\times d},
\qquad
M_Z=\mathcal F_{\mathrm{gate}}(Z,C_0).
\label{eq:ecsd_zero}
\end{equation}
With $C=C_0$, the second transformed branch reduces to $\tanh(\mathbf 1_Nb_h^{\top})$. Hence, $W_h$ and the $C$-dependent columns of $W_g$ do not affect the forward map, whereas $b_h$ and the $Z$-dependent fusion parameters remain trainable. Thus, $Z$ is the sole sample-dependent input used to construct $M_Z$. We call this fixed-interface configuration ECSD-Zero; $C_0$ is a parameter-free compatibility input, not a learned alignment mechanism or an independent contribution.

The scene Transformer attends to $M_Z$ and models
\begin{equation}
p_{\mathrm{scene}}(Y\mid M_Z)
=
\prod_{t=1}^{L}
p_{\mathrm{scene}}(y_t\mid y_{<t},M_Z),
\label{eq:objective}
\end{equation}
where $y_{<t}$ denotes the preceding scene-caption tokens. During CE training, both decoders use teacher forcing: $Z$ supplies the auxiliary phrase and diversity losses, $M_Z$ supplies the primary scene-caption loss, and $F_A$ supplies the binary change loss.
At inference, the local path is omitted:
\begin{equation}
\begin{aligned}
(I_1,I_2)&\rightarrow(F_1,F_2)\rightarrow F_A,\\
F_A&\rightarrow Z\rightarrow
\mathcal F_{\mathrm{gate}}(Z,\mathbf 0)\rightarrow M_Z,\\
M_Z&\rightarrow D_{\mathrm{scene}}\rightarrow\widetilde Y,\\
F_A&\rightarrow s_{\mathrm{chg}}\rightarrow p_{\mathrm{chg}},\\
(p_{\mathrm{chg}},\widetilde Y)&
\xrightarrow{\text{caption gate}}\widehat Y.
\end{aligned}
\label{eq:inference_graph}
\end{equation}
Table~\ref{tab:component_use} summarizes the component schedule. The auxiliary phrase decoder is a cross-entropy (CE) training head only; local hidden states are never supplied to the scene decoder.
\begin{table}[t]
\centering
\caption{Use of HIMEC components during CE training, SCST, and inference.}
\label{tab:component_use}
\footnotesize
\setlength{\tabcolsep}{3pt}
\begin{tabular}{@{}p{0.47\linewidth}ccc@{}}
\toprule
Component & CE train & SCST train & Inference \\
\midrule
Encoder, DCR, and change-query encoder & Yes & Yes & Yes \\
Scene decoder & Yes & Yes & Yes \\
Binary change head & Yes & No & Yes \\
Caption gate & No & No & Yes \\
Auxiliary local phrase decoder & Yes & No & No \\
Local hidden states supplied to scene decoder & No & No & No \\
Fixed zero second input & Yes & Yes & Yes \\
\bottomrule
\end{tabular}
\end{table}
The final returned caption is
\begin{equation}
\widehat Y=
\begin{cases}
Y_{\mathrm{nc}},
& p_{\mathrm{chg}}<\tau,\\
\widetilde Y,
& p_{\mathrm{chg}}\geq\tau,
\end{cases}
\label{eq:final_caption}
\end{equation}
where $Y_{\mathrm{nc}}$ is the fixed no-change sentence and $\tau$ is the applicable threshold specified in Section~\ref{sec:metrics}.

\subsection{Training Objective}
\label{sec:losses}

\subsubsection{Stage 1: Cross-Entropy Training}

During Stage 1, the model minimizes
\begin{equation}
\mathcal L_{\mathrm{CE}}
=
\lambda_g\mathcal L_{\mathrm{scene}}
+\lambda_l\mathcal L_{\mathrm{local}}
+\lambda_d\mathcal L_{\mathrm{div}}
+\lambda_c\mathcal L_{\mathrm{chg}}.
\label{eq:total_loss}
\end{equation}
The scene and local terms use token-level cross-entropy with label smoothing $\epsilon=0.1$. Let $\Omega_{\mathrm{scene}}$ denote the nonpadding next-token positions in the selected reference captions. $\mathcal L_{\mathrm{scene}}$ is $\operatorname{CE}_{\epsilon}(\ell^{\mathrm{scene}}_{bt},y_{bt})$ averaged over $(b,t)\in\Omega_{\mathrm{scene}}$, where $\ell^{\mathrm{scene}}_{bt}$ is the scene-decoder logit vector. For the local loss, let $\mathcal V$ be the set of valid phrase slots from changed samples and let $\Omega_{bk}$ contain the nonpadding next-token positions for slot $k$ of sample $b$. The implemented slot-balanced loss is
\begin{equation}
\mathcal L_{\mathrm{local}}
=
\frac{1}{|\mathcal V|}
\sum_{(b,k)\in\mathcal V}
\frac{1}{|\Omega_{bk}|}
\sum_{t\in\Omega_{bk}}
\operatorname{CE}_{\epsilon}
\left(
\ell_{bkt},e_{bkt}
\right),
\label{eq:local_loss}
\end{equation}
where $\ell_{bkt}$ is the local-decoder logit vector. The loss is set to zero when a batch contains no valid phrase slots. It propagates through $z_k$ to the query encoder, DCR, and unfrozen encoder parameters.

Let $\mathcal B_{+}=\{b:c_b=1\}$ and $\widehat z_{bk}=z_{bk}/\max(\lVert z_{bk}\rVert_2,\delta)$, where $\delta=10^{-12}$. For $\mathcal B_{+}\neq\varnothing$, the diversity term is
\begin{equation}
\mathcal L_{\mathrm{div}}
=
\frac{1}{|\mathcal B_{+}|}
\sum_{b\in\mathcal B_{+}}
\frac{2}{N(N-1)}
\sum_{1\le i<j\le N}
\widehat z_{bi}^{\top}\widehat z_{bj}.
\label{eq:div_loss}
\end{equation}
It is set to zero when $\mathcal B_{+}$ is empty and otherwise discourages the change-query tokens from collapsing to similar representations. The change loss is
\begin{equation}
\mathcal L_{\mathrm{chg}}
=
\operatorname{BCE}_{w_{+}}
\left(\boldsymbol{s}_{\mathrm{chg}},\boldsymbol{c}\right),
\label{eq:change_loss}
\end{equation}
where $\operatorname{BCE}_{w_{+}}$ is binary cross-entropy averaged over the batch, $\boldsymbol{c}$ contains the image-level change labels, and positive samples are weighted by the negative-to-positive ratio $w_{+}$ of the training split.

We use $\lambda_g=1.0$, $\lambda_l=0.3$, $\lambda_d=0.05$, and $\lambda_c=0.5$ for LEVIR-CC and SECOND-CC. $\mathcal L_{\mathrm{scene}}$ updates the scene decoder, gated fusion block, query encoder, DCR, and unfrozen encoder parameters. $\mathcal L_{\mathrm{local}}$ updates the local-specific decoder parameters, the token embedding shared with the scene decoder, and the upstream representation path through $Z$. $\mathcal L_{\mathrm{div}}$ acts on $Z$ and its upstream modules. $\mathcal L_{\mathrm{chg}}$ updates the change head and the shared visual path before query-based change encoding. Under ECSD-Zero, no scene-loss gradient or feature path passes through local-decoder hidden states.

\subsubsection{Stage 2: Self-Critical Sequence Training}

The scene-generation path is subsequently fine-tuned with SCST~\cite{rennie2017scst}. Let $\Theta$ collect its trainable parameters, let $Y^s\sim p_{\Theta}(\cdot\mid I_1,I_2)$, and let $Y^g$ be the greedily decoded caption. The SCST objective is
\begin{equation}
\mathcal L_{\mathrm{SCST}}(\Theta)
=
-\mathbb E_{Y^s\sim p_{\Theta}}
\left[r_{\mathcal B}(Y^s)\right],
\label{eq:scst}
\end{equation}
with the sampled policy gradient
\begin{equation}
\nabla_\Theta\mathcal L_{\mathrm{SCST}}
\approx
-\operatorname{sg}\!\left[
r_{\mathcal B}(Y^s)-r_{\mathcal B}(Y^g)
\right]
\nabla_\Theta\log p_\Theta(Y^s).
\label{eq:scst_grad}
\end{equation}
The greedy reward is the self-critical baseline. In the implementation, $r_{\mathcal B}$ is a mini-batch, CIDEr-D-inspired $n$-gram-overlap reward on model-vocabulary tokens. It uses all references for each image and estimates document frequencies from the current mini-batch; it is distinct from the corpus-level CIDEr-D evaluator used for reporting.

The scene decoder is updated with the main SCST learning rate, while the encoder, DCR, and query-encoding modules participating in scene generation use a lower learning rate. The SCST forward pass omits the local phrase branch and the change head. Consequently, the local-specific decoder parameters and change head receive no SCST gradients; the shared token embedding is updated only through the scene-generation path, and no auxiliary phrase loss is applied.

\section{Experiments}
\label{sec:setup}

\subsection{Datasets}
\label{sec:datasets}

\begin{table}[!t]
\centering
\caption{Statistics of the three RSICC benchmarks used in the experiments.}
\label{tab:datasets}
\tstrut
\setlength{\tabcolsep}{3.5pt}
\footnotesize
\begin{tabular}{lccc}
\toprule
\textbf{Property} & \textbf{LEVIR-CC} & \textbf{SECOND-CC} & \textbf{DUBAI-CC} \\
\midrule
Pairs (train/val/test)& 6815/1333/1929 & 4219/595/1227  & 300/50/150 \\
Captions per pair     & 5              & 2 to 5$^\dagger$ & 5 \\
Changed pairs (\%)    & 50.0           & 71.7             & 65.0 \\
Mean length (tokens)  & $11.0 \pm 3.7$ & $12.6 \pm 3.7$   & $8.8 \pm 3.0$ \\
Maximum length        & 39             & 21               & 25 \\
Vocabulary size       & 860            & 956               & 298 \\
\bottomrule
\end{tabular}
\par\vspace{1pt}
{\footnotesize $^\dagger$SECOND-CC's official augmented release adds one offline transformed view for each training and validation pair, producing 8438/1190/1227 indexed samples in our loader. It is designed for five captions per unique pair; in the supplied JSON, 31 pairs contain two to four captions. DUBAI-CC tiles ($50{\times}50$ pixels) are resized to $256{\times}256$ for training.\par}
\end{table}

We evaluate HIMEC on three public RSICC datasets summarized in Table~\ref{tab:datasets}. LEVIR-CC~\cite{liu2022rsicformer} contains 10,077 co-registered bitemporal pairs with five captions per pair and is the primary benchmark. SECOND-CC~\cite{karaca2025mmodalcc,chen2023second} contains 6041 unique pairs split into 4219 training, 595 validation, and 1227 test pairs. We use its official augmented JSON release, which provides one offline transformed view for each training and validation pair; these views increase the corresponding loader sample counts without increasing the number of unique pairs. SECOND-CC differs in scene distribution, annotation style, and change characteristics and supports replication on a second benchmark after separate training. DUBAI-CC~\cite{hoxha2022change} contains 500 low-resolution pairs, including 300 training pairs, and is used to assess path removal under data scarcity (Section~\ref{sec:causal_control}). Because its public caption JSON does not provide binary change flags, we derive them from the reference captions: a pair is marked unchanged when at least one reference contains a predefined no-change marker. These automatically derived labels are used only for the DUBAI-CC change-head diagnostic.

The datasets provide no human-audited phrase counts. We therefore omit parser-derived counts from Table~\ref{tab:datasets} and report them only as a post-hoc diagnostic in Section~\ref{sec:phrase_subset}. The $N$-query interface represents capacity for several latent change summaries; individual queries are not assumed to correspond to distinct semantic changes.

\subsection{Evaluation Protocol}
\label{sec:metrics}

Caption quality is measured with BLEU-4~\cite{papineni2002bleu}, METEOR~\cite{banerjee2005meteor}, ROUGE-L~\cite{lin2004rouge}, and CIDEr-D~\cite{vedantam2015cider}. CIDEr-D is the primary captioning metric and is abbreviated as CIDEr in prose and compact tables. Change-F1 (CF1) evaluates the binary change head by thresholding its sigmoid probability with the threshold specified below and comparing the resulting change/no-change decision with the dataset label.

\textbf{Scoring.} Unless a table is explicitly marked as a legacy-scorer analysis, experiments use \texttt{pycocoevalcap} v1.2~\cite{chen2015cococap} with Penn Treebank (PTB) tokenization through \texttt{PTBTokenizer} for candidate and reference captions. CIDEr-D uses four-gram matching $(n=4)$ and $\sigma=6.0$; METEOR uses the Java METEOR 1.5 implementation.

\textbf{Inference.} Beam search uses width 3, a length-penalty exponent of 0.7, trigram blocking, and a repetition penalty of 1.2. LEVIR-CC uses maximum and minimum changed-caption lengths of 40 and 8 tokens. SECOND-CC uses 40 and 5 tokens, and DUBAI-CC uses 25 and 3 tokens. For cross-entropy and diagnostic checkpoints, the change threshold is selected on the validation split from 0.30 to 0.70 at intervals of 0.05 on LEVIR-CC and SECOND-CC and 0.10 on DUBAI-CC. The final SCST checkpoints use $\tau=0.50$ because the validation-selected cross-entropy threshold was not propagated through SCST checkpointing. Samples below the applicable threshold receive the fixed caption ``there is no difference.''

\textbf{Seeds and statistics.} Three-seed aggregates identified in the tables use seeds 42, 123, and 2024; all other rows state their seed or run count explicitly. Statistical comparisons use a two-sided paired bootstrap over test pairs with 10,000 resamples. For three-seed comparisons, per-image CIDEr-D contributions are averaged across seeds before resampling. These image-level tests are conditional on the trained checkpoints and do not incorporate training-seed uncertainty. A small conditional $p$-value cannot correct differences in training configuration or checkpoint provenance. We report $p$-values and 95\% confidence intervals where appropriate, and a nonsignificant result is not treated as evidence of equivalence.

Unless stated otherwise, this protocol is used for all datasets. The legacy-scorer results in Tables~\ref{tab:mismatch} and~\ref{tab:robust} are explicitly marked exceptions.

\subsection{Implementation Details}
\label{sec:implementation}

Experiments are implemented in PyTorch and run on one NVIDIA RTX 4070 Laptop GPU with 8\,GB memory. Images are resized to $256{\times}256$, normalized with ImageNet statistics, and synchronously flipped horizontally and vertically during cross-entropy training. The vocabulary includes tokens occurring at least once. Scene/local maximum lengths are 40/16 tokens on LEVIR-CC, 40/22 on SECOND-CC, and 25/16 on DUBAI-CC.

Training comprises 40 cross-entropy epochs with batch size 16 and 10 SCST epochs with batch size 8. Cross-entropy training uses AdamW, weight decay $10^{-4}$, five epochs of linear warmup, cosine decay, and gradient clipping at 0.5. Cross-entropy checkpoints are selected with the configured validation composite, $0.7$ CIDEr-D plus $0.3$ Change-F1, whereas SCST checkpoints are selected by validation CIDEr-D. For the final Swin-T DCR model, the nonencoder and encoder learning rates are $10^{-4}$ and $3{\times}10^{-5}$; diagnostic models without DCR use $2{\times}10^{-4}$ and $5{\times}10^{-5}$. During SCST, the scene decoder learning rate is $5{\times}10^{-5}$, and the remaining participating modules use $5{\times}10^{-6}$. The ECSD-Zero SCST seeds were initialized from a common cross-entropy checkpoint, whereas DCR-SCST used seed-specific cross-entropy checkpoints; the two groups also inherit the different cross-entropy learning rates above. Their comparison is therefore not a one-factor DCR ablation.

Swin-T~\cite{liu2021swin} is the default backbone, and ResNet-50~\cite{he2016resnet} is evaluated as an alternative. Both are initialized with ImageNet weights. The local and scene decoders contain two and three Transformer layers~\cite{vaswani2017attention}, respectively. Both use eight attention heads, a hidden dimension of 512, and dropout 0.1.

\subsection{Diagnostic Cascaded Baseline}
\label{sec:cascaded_method}

This subsection defines the local-to-scene cascade used only for the conditioning diagnosis. It shares the visual and query representations with the experimental implementation but differs from adopted HIMEC by supplying local-decoder states to the scene decoder. It is therefore a comparison architecture, not a component of the adopted inference graph in~\eqref{eq:inference_graph}.

Let $u_{k,t}^{r}$ be the final-layer local-decoder state at valid position $t$ for phrase slot $k$, where $r\in\{\mathrm{TF},\mathrm{AR}\}$ denotes teacher-forced or autoregressive local decoding. The local representation is the projected mean of the valid token states:
\begin{equation}
h_k^{r}
=W_p\left(
\frac{1}{T_k^{r}}
\sum_{t=1}^{T_k^{r}}u_{k,t}^{r}
\right)+b_p,
\qquad
H^{r}=\{h_k^{r}\}_{k=1}^{N}.
\label{eq:local_state_method}
\end{equation}
The diagnostic baseline supplies $H^{\mathrm{TF}}$ to the scene decoder during training and $H^{\mathrm{AR}}$ during inference; we denote this teacher-forced/autoregressive schedule as TF/AR. Its scene memories are
\begin{equation}
M_{\mathrm{base}}^{\mathrm{train}}
=\mathcal F_{\mathrm{gate}}(Z,H^{\mathrm{TF}}),
\qquad
M_{\mathrm{base}}^{\mathrm{test}}
=\mathcal F_{\mathrm{gate}}(Z,H^{\mathrm{AR}}).
\label{eq:cascaded_memory_method}
\end{equation}
The different local-decoding regimes create the discrepancy quantified below. By contrast, adopted HIMEC always supplies the fixed $C_0$ in~\eqref{eq:ecsd_zero} and never transfers $H^r$.

\subsection{Quantifying the Inter-Decoder Hidden-State Discrepancy}
\label{sec:shift_measure}

\begin{figure}[t]
\centering
\includegraphics[width=0.95\linewidth,keepaspectratio]{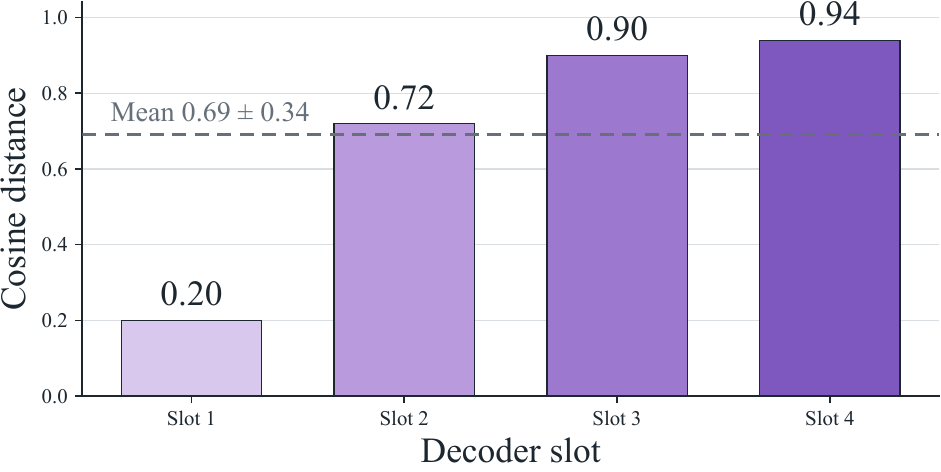}
\caption{Slotwise cosine distance between paired teacher-forced and autoregressive local states on changed LEVIR-CC validation pairs with valid phrase targets (Swin-T). The dashed line marks the overall mean of 0.69.}
\label{fig:mismatch}
\end{figure}

For the cascaded baseline in Section~\ref{sec:cascaded_method}, the slotwise discrepancy between the pooled states $h_k^{\mathrm{TF}}$ and $h_k^{\mathrm{AR}}$ is
\begin{equation}
d_k
=
1-\cos\!\left(
h_k^{\mathrm{TF}},
h_k^{\mathrm{AR}}
\right),
\label{eq:cosine_dist}
\end{equation}
where larger values indicate greater disagreement. The states and their use by the scene decoder are defined in \eqref{eq:local_state_method} and~\eqref{eq:cascaded_memory_method}. We compute $d_k$ with Swin-T on changed LEVIR-CC validation pairs with valid local supervision; Fig.~\ref{fig:mismatch} reports the result. This diagnostic is not part of the adopted HIMEC inference graph.

The mean distance is $0.69 \pm 0.34$, with $(d_1,\ldots,d_4)=(0.20,0.72,0.90,0.94)$. Distances increase in the later, less frequently supervised slots. ResNet-50 gives a similar mean of $0.69 \pm 0.30$.

\subsubsection{Replication Across Decoder Families}

Replacing the local Transformer with a comparably sized long short-term memory (LSTM) decoder and retraining from scratch yields 134.4 CIDEr and a mean distance of $0.69 \pm 0.25$, with $(d_1,\ldots,d_4)=(0.34,0.70,0.85,0.88)$. The discrepancy therefore persists across the evaluated local-decoder families.

\subsubsection{Reproduction in an Independent Architecture}

We also construct an independent cascaded local-to-scene model with a ResNet-50 bilateral encoder, a local LSTM decoder, and an attention-based LSTM scene decoder, without DCR or query-based change encoding. On LEVIR-CC, teacher-forced conditioning yields a cosine distance of 0.73 and a test CIDEr of 39.7. After retraining with the downstream local-state path zeroed, the unused local representations have a distance of 0.33, while the actual scene-decoder input discrepancy is zero by construction; CIDEr rises to 115.5. On SECOND-CC, the corresponding teacher-forced local-state distance is 0.95, and CIDEr increases from 27.8 to 78.6 after path zeroing. These results are mechanism checks and are not used for performance comparison with HIMEC.

\subsection{Controlled Analysis of the Inter-Decoder Mismatch}
\label{sec:causal_control}

The controls separate conditioning consistency from image-specific local-state content. LEVIR-CC is the primary test, with replication on SECOND-CC and a data-scarce test on DUBAI-CC.

\subsubsection{Controlled Comparison Protocols}
\label{sec:interventions}

\begin{figure*}[!t]
\centering
\includegraphics[width=0.92\textwidth,keepaspectratio]{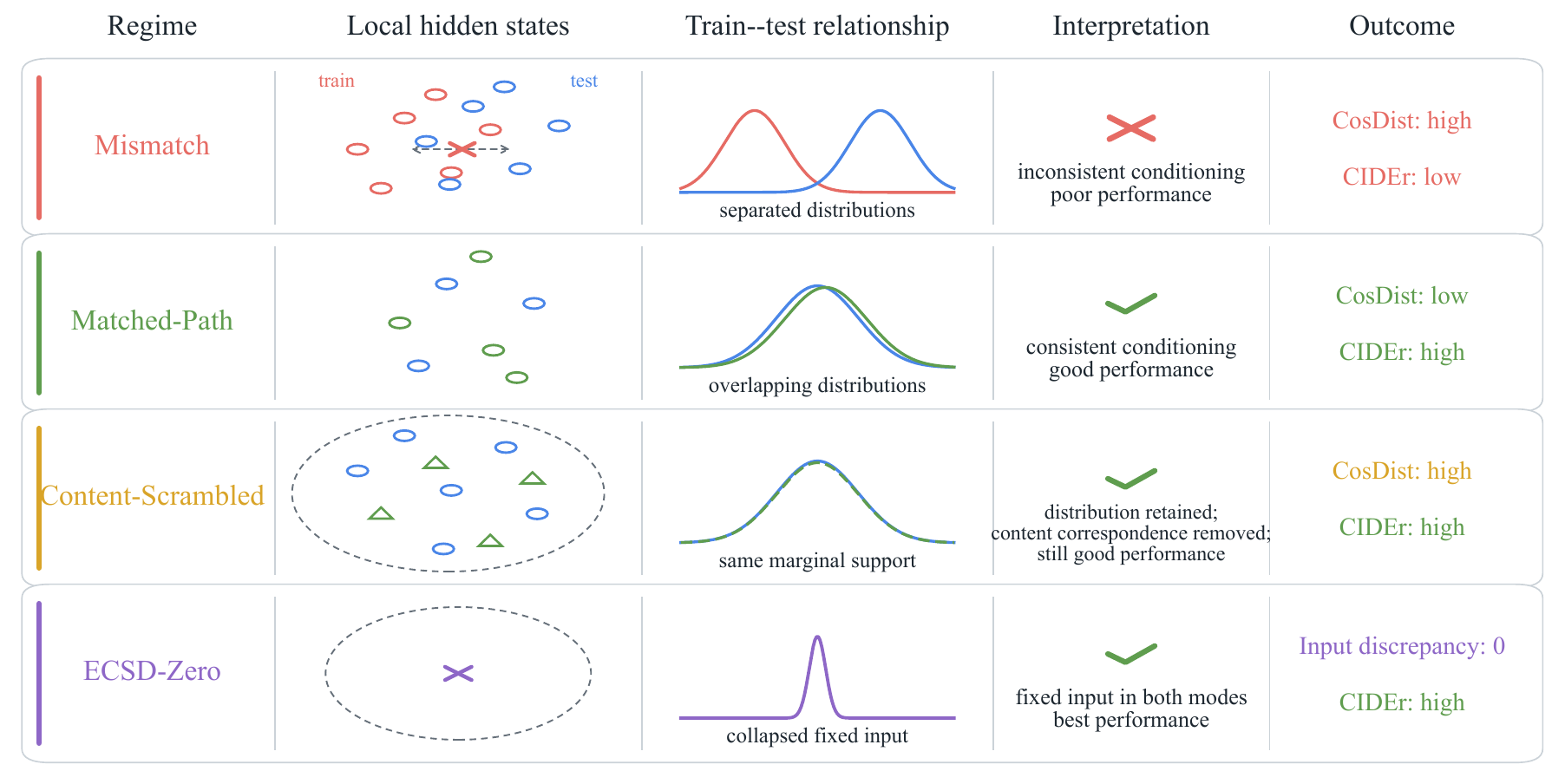}
\caption{Diagnostic conditioning regimes. TF/AR supplies teacher-forced local states during training and autoregressive states at inference. Matched-Path uses autoregressive states in both modes; Content-Scrambled permutes those states across samples. ECSD-Zero supplies a constant zero input and is HIMEC's adopted interface. CosDist denotes cosine distance for nonzero local states; the zero-input regime has zero train--inference input discrepancy.}
\label{fig:regimes}
\end{figure*}

The \emph{regime-matched} control, reported as Matched-Path, supplies detached autoregressive local states to the scene decoder during both training and inference. The \emph{content-permuted} control, reported as Content-Scrambled, additionally permutes these states across samples within each mini-batch, preserving their marginal distribution while removing image-state correspondence. The \texttt{no\_local} control removes the local decoder and its auxiliary loss. Greedy-Aligned uses autoregressive states during training while retaining the scene-loss gradient path. ECSD-Gate learns to suppress the second input, whereas ECSD-Consistency adds a teacher-forced/autoregressive representation-alignment loss. Fig.~\ref{fig:regimes} summarizes the principal conditioning regimes.

\subsubsection{Comparison with Exposure-Bias Remedies}
\label{sec:mismatch_ablation}

The legacy within-scorer values are moved to Appendix Table~\ref{tab:mismatch} to prevent comparison with the standard PTBTokenizer results. In that diagnostic table, annealed scheduled sampling and professor forcing score below the cascaded baseline. Fixed-rate scheduled sampling reaches 131.03 CIDEr at $p=0.75$ but remains 5.56 points below ECSD-Zero. Matched-Path, Content-Scrambled, and ECSD-Zero have higher observed scores than the interpolation methods under the same legacy scorer.

\subsubsection{Regime-Matched Control}
Under the three-seed PTB evaluation, Matched-Path reaches $135.25\pm0.28$ CIDEr. In the same-seed legacy comparison in Table~\ref{tab:mismatch}, it gains 12.65 points over the cascaded baseline and recovers 85.5\% of the ECSD-Zero improvement. Retaining the path while matching its autoregressive regime therefore recovers most of the lost performance. Greedy-Aligned reaches 133.36 CIDEr under the same legacy scorer but remains below the detached control.

\subsubsection{Content-Permuted Control}

Content-Scrambled reaches a three-seed PTB mean of $134.97\pm0.98$ CIDEr, compared with $135.25\pm0.28$ for Matched-Path. The paired image bootstrap detects no difference $(p=0.63$, 95\% confidence interval $[-1.39,0.85])$, so this experiment identifies no penalty from removing image-state correspondence but does not establish equivalence. In the corrected same-backbone comparison, Swin-T ECSD-Zero and complete local-branch removal obtain three-seed means of $135.62\pm1.08$ and $136.34\pm0.14$, respectively. This descriptive difference does not establish a captioning benefit from auxiliary local supervision. Within the evaluated interfaces, these controls provide no evidence that sample-dependent local states are useful; they do not imply that ECSD-Zero learns an alignment transformation.

\subsubsection{Replication on SECOND-CC}

On SECOND-CC, the cascaded baseline obtains 60.77 CIDEr, ECSD-Zero obtains $75.67\pm1.46$, and Matched-Path obtains $76.99\pm0.78$. The Matched-Path difference relative to ECSD-Zero is $+1.32$, with a 95\% confidence interval of $[-0.36,+2.96]$ and $p=0.117$. Both consistent-interface variants outperform the mismatched baseline. The interval for Matched-Path minus ECSD-Zero includes zero, so their difference is not resolved.

\subsubsection{Stress Test on DUBAI-CC}

On DUBAI-CC, under the explicitly legacy scorer, the three-seed CIDEr means are $45.53\pm3.78$ for the cascaded baseline, $48.03\pm2.42$ for Matched-Path, and $62.61\pm7.67$ for ECSD-Zero. The corresponding Change-F1 means are $0.87\pm0.02$, $0.89\pm0.02$, and $0.87\pm0.03$. This 300-pair training set does not reproduce the matched-path recovery.

To test whether sample count alone explains that result, we draw a fixed 300-pair subset from the LEVIR-CC training split (subset seed 777) and retrain the three interfaces with seed 42, identical optimization, and the standard PTBTokenizer evaluation. TF/AR, Matched-Path, and ECSD-Zero obtain 110.22, 121.45, and 123.77 CIDEr, respectively; their BLEU-4 scores are 25.75, 37.39, and 38.33. Under the same nominal training-set size, regime matching recovers 11.23 CIDEr, and ECSD-Zero gains 13.55 CIDEr over TF/AR. This single-subset, single-seed control does not reproduce the DUBAI-CC failure, so that failure cannot be attributed to sample count alone within the evaluated settings. The experiment does not isolate whether the remaining difference arises from spatial resolution, change distribution, annotation style, or optimization variance.

\subsection{Overall Performance}
\label{sec:results}
\label{sec:main_results}

\begin{table*}[t]
\centering
\caption{Reported results on the LEVIR-CC test set. Evaluation protocols and auxiliary inputs differ; cross-method values are descriptive.}
\label{tab:main}
\tstrut
\setlength{\tabcolsep}{4.2pt}
\footnotesize
\begin{tabular}{@{}llS[table-format=2.2]S[table-format=2.2]
                   S[table-format=2.2]S[table-format=3.2]S[table-format=1.4]@{}}
\toprule
\textbf{Method} & \textbf{Architecture} & {\textbf{BLEU-4}} & {\textbf{METEOR}} &
 {\textbf{ROUGE-L}} & {\textbf{CIDEr-D}} & {\textbf{CF1}} \\
\midrule
DUDA~\cite{cai2021duda}                 & CNN$+$LSTM        & 57.79 & 34.21 & 69.18 & 124.32 & 0.9120 \\
MCCFormer-D~\cite{liu2022mccformer}     & CNN$+$Transformer & 56.38 & 32.91 & 68.56 & 124.44 & 0.9135 \\
RSICCformer~\cite{liu2022rsicformer}    & CNN$+$Transformer & 62.77 & 37.84 & 73.45 & 134.12 & 0.9245 \\
Chg2Cap~\cite{chang2023chg2cap}         & CNN$+$Transformer & 64.39 & 38.75 & 74.92 & 136.61 & 0.9305 \\
SAT-Cap~\cite{yao2025satcap}            & CNN$+$Transformer & 65.82 & 39.45 & 75.82 & 140.23 & 0.9365 \\
DSPG~\cite{li2026dspg}                  & CNN$+$Transformer & 66.71 & 41.61 & 76.81 & 143.34 & NR \\
Semantic-CC~\cite{semanticcc2024}       & SAM$+$LLM         & 64.51 & 40.58 & 77.76 & 138.51 & NR \\
RSCaMa~\cite{rscama2024}                & State space       & 65.24 & 39.91 & 75.24 & 136.56 & NR \\
CCExpert-7B~\cite{ccexpert2024}         & MLLM (7B)         & 65.49 & 41.82 & 76.55 & 143.32 & NR \\
\midrule
HIMEC (ours)                             & RGB-only change-query Transformer & 49.01 & 35.10 & 74.32 & 142.81 & \textbf{0.9450} \\
\bottomrule
\end{tabular}
\par\vspace{1pt}
{\footnotesize HIMEC uses the protocol in Section~\ref{sec:metrics}; its three-seed means are $142.81\pm0.60$ CIDEr-D and $0.9450\pm0.0021$ CF1. Other entries are published values. Published CF1 values are included when available and marked NR otherwise; because the external models were not re-evaluated under our change-label pipeline, this comparison is descriptive. Bold marks HIMEC's highest reported CF1; competing values are left unformatted. SAM: Segment Anything Model; LLM: large language model; MLLM: multimodal large language model; NR: not reported.\par}
\end{table*}

HIMEC obtains $142.81\pm0.60$ CIDEr over seeds 42, 123, and 2024, with individual scores of 143.40, 143.03, and 141.99. The matched direct-memory control reaches $139.51\pm3.40$ CIDEr. The 3.30-point mean difference favors the structured query interface under the evaluated training protocol; across these three runs, its standard deviation is 0.60 rather than 3.40. Change-F1 also increases from $0.9393\pm0.0053$ for direct memory to $0.9450\pm0.0021$ for change-query memory. Both modes contain 46.23 million parameters, so the score difference is not accompanied by a parameter-count advantage. For external context, published CIDEr values include 143.34 for DSPG, 143.32 for CCExpert-7B, and 140.23 for SAT-Cap. Because the evaluation pipelines differ, we make no ranking or significance claim.

\subsubsection{Comparison with Recent Methods}
\label{sec:latest_comparison}

DSPG uses a CNN--Transformer architecture, whereas CCExpert-7B is a 7-billion-parameter MLLM. MModalCC reports 144.89 CIDEr on the separate LEVIR-MCI benchmark with segmentation maps at inference; this value is excluded from Table~\ref{tab:main} because its inputs and benchmark differ.

HIMEC has a lower BLEU-4 score than several recent methods (49.01 versus 62.8 to 66.7). Under the same PTBTokenizer pipeline, RSICCformer and Chg2Cap obtain CIDEr scores of 134.19 and 136.63, within 0.1 of their published values, and BLEU-4 scores of 64.88 and 64.39. The BLEU-4 gap therefore remains under a common scorer and cannot be dismissed as a scoring-pipeline artifact.

We decompose the seed-42 output with a lowercase word tokenizer. On the 964 changed pairs, candidate and mean-reference lengths are 9.54 and 10.93 words, but the corpus brevity penalty is 0.993; length alone therefore accounts for little of the observed reduction. Clipped 1--4-gram precisions decline from 0.758 to 0.501, 0.294, and 0.158. Only 11.5\% of changed captions are unique, and the three most frequent changed-caption templates cover 38.3\% of the subset. The lower BLEU-4 is thus associated primarily with limited high-order phrase overlap and template concentration, in addition to modest underlength. The five references also show substantial wording variation (mean pairwise 4-gram Jaccard 0.009 under this diagnostic tokenizer). CIDEr can still reward agreement on relatively informative content $n$-grams through consensus-weighted TF--IDF, which is consistent with the smaller CIDEr gap. This analysis characterizes the output but does not establish that ECSD-Zero causes the linguistic pattern; public predictions for the recent comparison methods are unavailable for the same decomposition.

Rescoring the released MModalCC checkpoint on SECOND-CC reproduces its reported CIDEr score in Table~\ref{tab:second}. The released RSICCformer and Chg2Cap predictions are also rescored under our PTBTokenizer pipeline; the remaining cross-method entries use published values.

\subsection{Controlled and Exploratory Component Analysis}
\label{sec:ablation_study}

\subsubsection{Configuration-Matched Component Ablations}
\label{sec:ablation}

\begin{table}[!t]
\centering
\caption{Single-seed, configuration-matched cross-entropy component ablations on LEVIR-CC.}
\label{tab:component}
\tstrut
\setlength{\tabcolsep}{3.2pt}
\scriptsize
\begin{tabular}{@{}lS[table-format=2.2]S[table-format=3.2]
S[table-format=+2.2]S[table-format=1.4]@{}}
\toprule
\textbf{Variant} & {\textbf{B-4}} & {\textbf{CIDEr}} &
{\textbf{$\Delta$C}} & {\textbf{CF1}} \\
\midrule
Full HIMEC (CE)        & 41.57 & 137.65 & {0.00}  & 0.9435 \\
$-\mathcal{L}_{\mathrm{div}}$ & 42.55 & 138.28 & {+0.63} &
\multicolumn{1}{c}{\underline{0.9471}} \\
Repeated pooled $F_A$  &
\multicolumn{1}{c}{\textbf{44.51}} &
\multicolumn{1}{c}{\textbf{140.82}} & {+3.17} & 0.9467 \\
$-$difference interaction & 35.29 & 121.49 & {-16.16} & 0.8825 \\
$-$phrase supervision  &
\multicolumn{1}{c}{\underline{42.76}} &
\multicolumn{1}{c}{\underline{138.79}} & {+1.14} &
\multicolumn{1}{c}{\textbf{0.9481}} \\
$-$appearance stream   & 40.84 & 134.66 & {-2.99} & 0.9328 \\
$-$disappearance stream& 42.42 & 138.69 & {+1.04} & 0.9459 \\
$-$shared context      & 41.70 & 136.53 & {-1.12} & 0.9446 \\
\bottomrule
\end{tabular}
\par\vspace{1pt}
{\footnotesize All rows use Swin-T, ECSD-Zero, seed 42, 40 CE epochs, identical optimizer and learning rates, beam width 3, and PTBTokenizer. $\Delta$C is the CIDEr change relative to the full row; positive values mean that removal increased the observed score. ``Repeated pooled $F_A$'' replaces the two-layer query encoder and does not remove spatial supervision, which HIMEC does not use. Bold and underline mark the highest and second-highest observed values per metric; formatting does not imply significance.\par}
\end{table}

These ablations change one factor at a time. Removing the difference interaction causes the largest reduction ($-16.16$ CIDEr), while removing appearance and shared context reduces CIDEr by 2.99 and 1.12 points. These three interventions lower the observed score under this seed. Removing the diversity objective, phrase supervision, two-layer query encoder, or disappearance stream does not reduce the observed CE score, so we do not attribute an isolated gain to those components. The three-seed query-versus-direct comparison in Table~\ref{tab:seeds} evaluates the complete query-memory interface after SCST; it does not establish an independent contribution from the specific two-layer query encoder, whose replacement did not lower the single-seed CE score.

\subsubsection{Query-Count Sensitivity}
\label{sec:slot_count}

\begin{table}[!t]
\centering
\caption{Single-seed query-count sensitivity on LEVIR-CC using Swin-T.}
\label{tab:n_ablation}
\tstrut
\setlength{\tabcolsep}{4pt}
\footnotesize
\begin{tabular}{@{}cS[table-format=3.2]S[table-format=2.2]
                  S[table-format=2.2]S[table-format=2.2]S[table-format=1.4]@{}}
\toprule
$\boldsymbol{N}$ & {\textbf{CIDEr}} & {\textbf{B-4}} &
{\textbf{M}} & {\textbf{R-L}} & {\textbf{CF1}} \\
\midrule
2           & 137.76 & 40.93 & 34.17 & 72.44 & 0.9452 \\
4 (default) & 134.27 & 43.65 & 33.74 & 71.67 & 0.9266 \\
8           & 137.61 & 41.84 & 34.09 & 71.99 & 0.9511 \\
\bottomrule
\end{tabular}
\end{table}

Table~\ref{tab:n_ablation} shows that, with backbone and seed matched, $N=2$ and $N=8$ exceed the default $N=4$ run by 3.49 and 3.34 CIDEr points, respectively. Because each setting is represented by one training run, the experiment does not identify an optimal query count. Four queries remain the pre-established capacity of the reported system; this comparison neither identifies $N=4$ as optimal nor demonstrates improved coverage of descriptions containing several changes.

\subsubsection{Exploratory Parser-Derived Phrase Subsets}
\label{sec:phrase_subset}

We conduct a post-hoc diagnostic using the same rule-based phrase extractor employed for auxiliary targets. Each changed LEVIR-CC test pair is assigned the maximum number of nonpadding phrases extracted from its five references, yielding 411 one-phrase and 553 two-or-more-phrase pairs; the remaining 965 pairs are unchanged. Across the three final HIMEC checkpoints, the one-phrase subset obtains $36.03\pm2.89$ BLEU-4, $49.16\pm1.68$ ROUGE-L, and $68.94\pm3.24$ mean per-image CIDEr-D contribution. The corresponding two-or-more-phrase values are $35.51\pm0.26$, $50.41\pm0.87$, and $72.96\pm2.41$. CIDEr-D is scored once with full-test-corpus document frequencies before its per-image contributions are grouped.

These subset scores are descriptive and are not difficulty-normalized evidence of a multi-phrase benefit. Phrase counts disagree among the five references for 548 of 964 changed pairs (56.8\%), the parser-derived bins have not been human-audited, and the available runs do not provide a fully matched direct-caption or component-removal comparison for these subsets. We therefore use this analysis only to characterize the final model and make no claim of phrase--region grounding or causal superiority on multi-change scenes.

\begin{table}[!t]
\centering
\caption{Configuration-matched scene-memory comparison on LEVIR-CC.}
\label{tab:seeds}
\tstrut
\setlength{\tabcolsep}{2.5pt}
\scriptsize
\begin{tabular}{@{}lccc@{}}
\toprule
\textbf{Scene memory} & \textbf{B-4} & \textbf{CIDEr} & \textbf{CF1} \\
\midrule
Direct fused features $F_A$ &
$\underline{46.80}\pm2.70$ &
$\underline{139.51}\pm3.40$ &
$\underline{0.9393}\pm0.0053$ \\
Change-query tokens $Z$ &
$\mathbf{49.01}\pm0.62$ &
$\mathbf{142.81}\pm0.60$ &
$\mathbf{0.9450}\pm0.0021$ \\
\bottomrule
\end{tabular}
\par\vspace{1pt}
{\footnotesize Means $\pm$ population standard deviations over seeds 42, 123, and 2024. Both rows use the same Swin-T encoder, DCR, ECSD-Zero input, CE and SCST schedules, decoding settings, and standard PTBTokenizer scoring; only the scene memory is changed. Bold and underline mark the higher and lower matched means. B-4: BLEU-4.\par}
\end{table}

\subsection{Sensitivity and Evaluation on SECOND-CC}

\subsubsection{Sensitivity to Image Degradation}

We test brightness, contrast, Gaussian blur, and spatial translation by applying the same deterministic perturbation to both images. This experiment measures sensitivity to these conditions and is not a claim of general corruption robustness.

Brightness and translation change CIDEr by at most 0.5 points and leave Change-F1 stable (Appendix Table~\ref{tab:robust}). High contrast reduces CIDEr by 4.5 points with little change in Change-F1. Gaussian blur has the largest effect: at severity 3.0, CIDEr falls from 135.18 to 111.29 and Change-F1 from 0.9374 to 0.2457. Targeted augmentation raises the corresponding values to 130.72 and 0.8012, reducing but not eliminating the blur sensitivity. The table is placed in the appendix because this perturbation harness uses the legacy scorer.

\begin{table}[!t]
\centering
\caption{Descriptive LEVIR-CC results from available checkpoints, grouped by backbone.}
\label{tab:backbone}
\tstrut
\setlength{\tabcolsep}{5pt}
\footnotesize
\begin{tabular}{@{}llS[table-format=3.2]@{{\,\footnotesize$\pm$\,}}S[table-format=1.2]@{}}
\toprule
\textbf{Variant} & \textbf{Backbone} & \multicolumn{2}{c}{\textbf{CIDEr}} \\
\midrule
\multirow{2}{*}{ECSD-Zero CE} & ResNet-50 & \multicolumn{2}{c}{136.07$^{\dagger}$} \\
                               & Swin-T    & \multicolumn{2}{c}{136.91$^{\dagger}$} \\
\addlinespace[2pt]
\multirow{2}{*}{DCR CE}       & ResNet-50 & 135.40 & 0.89 \\
                               & Swin-T    & \multicolumn{2}{c}{137.65$^{\dagger}$} \\
\addlinespace[2pt]
\multirow{2}{*}{DCR-SCST}     & ResNet-50 & 139.40 & 1.09 \\
                               & Swin-T    & 142.81 & 0.60 \\
\bottomrule
\end{tabular}
\par\vspace{1pt}
{\footnotesize Unmarked entries are three-seed means $\pm$ population standard deviations over seeds 42, 123, and 2024. $^{\dagger}$Single run with seed 42. Training and checkpoint provenance is not fully matched across all rows; the table is descriptive.\par}
\end{table}

Table~\ref{tab:backbone} shows a same-seed ECSD-Zero CE difference of 0.84 CIDEr in favor of Swin-T. DCR CE differs by 2.25 points, although its Swin-T entry is a single run. For DCR-SCST, the PTB-consistent three-seed means differ by 3.41 points. These comparisons describe the saved runs but do not constitute a controlled statistical backbone study.

\subsubsection{SECOND-CC Evaluation}
\label{sec:second_cc}

The ECSD diagnostic architecture is trained from scratch on SECOND-CC without dataset-specific architectural changes. These experiments use ECSD-Zero without DCR, so they replicate the interface comparison on a second dataset rather than evaluate the complete DCR-SCST system. Its rows in Table~\ref{tab:second} use the standard protocol in Section~\ref{sec:metrics}. The published reference rows use different scoring or additional inputs and are included only for context.

SECOND-CC emphasizes agricultural and land-cover changes and has predominantly single-clause captions. ECSD-Zero obtains 76.88 CIDEr with seed 42, and the corresponding single-seed SCST checkpoint scores 81.47. In the seed-42 comparison in Table~\ref{tab:second_ablation}, removing the difference interaction changes CIDEr by $-2.67$ points, replacing learned query encoding with repeated pooled features changes it by $+1.51$, and removing local supervision changes it by $+0.28$. These single-seed results do not establish general component effects.

\begin{figure*}[!t]
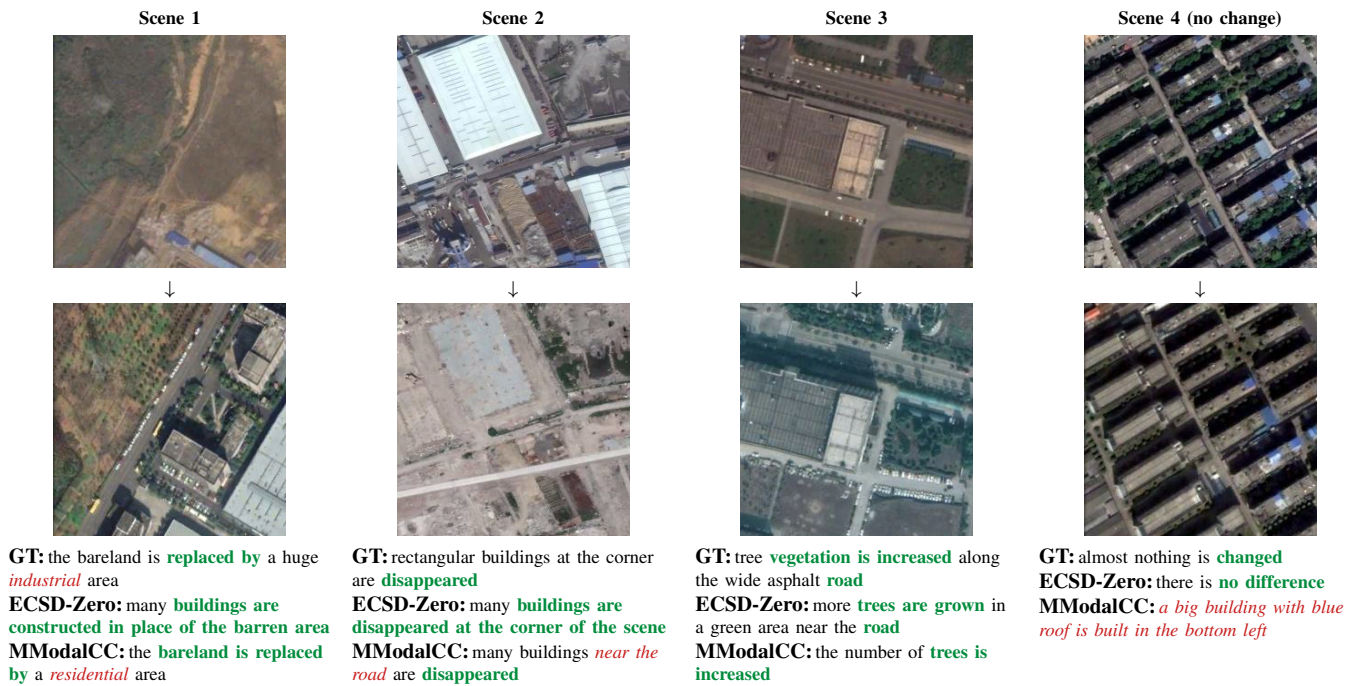

\centering
\setlength{\tabcolsep}{4pt}
\renewcommand{\arraystretch}{1.0}
\scriptsize
\newcommand{\sthumb}[1]{\includegraphics[width=\linewidth,height=\linewidth]{#1}}
\newcommand{\cml}[2]{{\footnotesize\textbf{#1:}}\,#2\par\vspace{1pt}}
\newcommand{\sscene}[6]{%
  \begin{minipage}[t]{\linewidth}\centering
    \begin{minipage}{0.72\linewidth}\centering
      \sthumb{#2}\\[2pt]$\downarrow$\\[2pt]\sthumb{#3}
    \end{minipage}\\[5pt]
    \raggedright\setlength{\parindent}{0pt}\linespread{1.05}\selectfont
    \cml{GT}{#4}\cml{ECSD-Zero}{#5}\cml{MModalCC}{#6}
  \end{minipage}}
\begin{tabular}{@{}*{4}{p{0.235\textwidth}}@{}}
  \centering\textbf{Scene 1} & \centering\textbf{Scene 2} &
  \centering\textbf{Scene 3} & \centering\textbf{Scene 4 (no change)}
  \tabularnewline[2pt]
  \sscene{}{compare/s972_A.png}{compare/s972_B.png}
    {the bareland is \cok{replaced by} a huge \cerr{industrial} area}
    {many \cok{buildings are constructed in place of the barren area}}
    {the \cok{bareland is replaced by} a \cerr{residential} area} &
  \sscene{}{compare/s282_A.png}{compare/s282_B.png}
    {rectangular buildings at the corner are \cok{disappeared}}
    {many \cok{buildings are disappeared at the corner of the scene}}
    {many buildings \cerr{near the road} are \cok{disappeared}} &
  \sscene{}{compare/s588_A.png}{compare/s588_B.png}
    {tree \cok{vegetation is increased} along the wide asphalt \cok{road}}
    {more \cok{trees are grown} in a green area near the \cok{road}}
    {the number of \cok{trees is increased}} &
  \sscene{}{compare/s84_A.png}{compare/s84_B.png}
    {almost nothing is \cok{changed}}
    {there is \cok{no difference}}
    {\cerr{a big building with blue roof is built in the bottom left}}
  \tabularnewline
\end{tabular}
\caption{Selected SECOND-CC captions from the ECSD-Zero diagnostic model and MModalCC. Phrases judged visually supported are shown in \cok{bold green}; errors or unsupported details are shown in \cerr{italic red}. Ground-truth captions are reproduced verbatim from the dataset.}
\label{fig:qual_second}
\end{figure*}

Fig.~\ref{fig:qual_second} compares the ECSD-Zero diagnostic model with the released MModalCC checkpoint under our qualitative harness. ECSD-Zero describes the building disappearance in Scene 2 and identifies Scene 4 as unchanged; MModalCC predicts a new building in Scene 4. These selected examples illustrate caption behavior but do not isolate a component effect.

\begin{table}[!t]
\centering
\caption{SECOND-CC captioning results for the diagnostic fixed-interface architecture, with published methods shown for context.}
\label{tab:second}
\tstrut
\setlength{\tabcolsep}{3.8pt}
\footnotesize
\begin{tabular}{@{}lccccc@{}}
\toprule
\textbf{Method} & \textbf{B-4} & \textbf{M} & \textbf{R-L} &
\textbf{C} & \textbf{CF1} \\
\midrule
RSICCformer$^{\dagger}$~\cite{liu2022rsicformer}
  & 34.00 & 26.20 & 54.70 & 81.00 & NE \\
MModalCC$^{\dagger}$~\cite{karaca2025mmodalcc}
  & 38.60 & 28.00 & 58.40 & 93.30 & NE \\
\midrule
Cascaded TF/AR$^{\ddagger}$ & 13.75 & \textbf{37.76} & 41.20 & 60.77 & \underline{0.9660} \\
ECSD-Zero (CE)$^{\ddagger}$ & \underline{20.02} & 20.24 &
\underline{50.59} & \underline{76.88} & \textbf{0.9671} \\
ECSD-Zero (SCST)$^{\ddagger}$ & \textbf{25.72} & \underline{20.74} &
\textbf{53.23} & \textbf{81.47} & 0.9460 \\
\bottomrule
\end{tabular}
\par\vspace{1pt}
{\footnotesize $^{\dagger}$Published results using beam width 4 and a custom scorer; MModalCC also uses semantic maps. Their CF1 values were not reported. $^{\ddagger}$Single run; the three-seed ECSD-Zero mean is $75.67\pm1.46$. Bold and underline rank only the three diagnostic rows; published context rows use a different protocol. The high TF/AR METEOR reflects exact matches on unchanged examples despite weak changed-caption scores. B-4: BLEU-4; M: METEOR; R-L: ROUGE-L; C: CIDEr-D; CF1: Change-F1; NE: not evaluated.\par}
\end{table}

\begin{table}[!t]
\centering
\caption{Single-seed, configuration-matched component comparison on SECOND-CC.}
\label{tab:second_ablation}
\tstrut
\setlength{\tabcolsep}{3.5pt}
\footnotesize
\begin{tabular}{@{}lccc@{}}
\toprule
\textbf{Variant} & \textbf{CIDEr} & \textbf{$\Delta$} &
\textbf{CF1} \\
\midrule
ECSD-Zero (reference) & 76.88 & Ref.    & 0.9671 \\
no diff. interaction  & 74.21 & $-2.67$ & 0.9605 \\
repeated pooled $F_A$    & 78.39 & $+1.51$ & 0.9654 \\
no local supv.         & 77.16 & $+0.28$ & 0.9672 \\
\bottomrule
\end{tabular}
\par\vspace{1pt}
{\footnotesize Results use ResNet-50, seed 42, and the PTBTokenizer pipeline. ``Repeated pooled $F_A$'' is the saved checkpoint named \texttt{no\_grounding}; no spatial grounding is removed. Only within-table changes are interpreted.\par}
\end{table}

\subsection{Efficiency Analysis}
\label{sec:efficiency}

HIMEC ECSD-Zero uses 42.8 million parameters, compared with 201.5 million for RSICCformer and 329.0 million for Chg2Cap, and has lower encoder GFLOPs (Table~\ref{tab:efficiency}). Its measured latency is much higher, but the non-equivalent decoding implementations make this an implementation rather than an architecture comparison. ECSD-Zero adds no separate inference step.

\begin{table}[!t]
\centering
\caption{Efficiency measurements for the evaluated checkpoints on one RTX 4070 GPU.}
\label{tab:efficiency}
\tstrut
\setlength{\tabcolsep}{3pt}
\footnotesize
\begin{tabular}{@{}lS[table-format=3.1]S[table-format=2.1]
                   S[table-format=3.0]S[table-format=3.2]@{}}
\toprule
\textbf{Model} & \textbf{Params} & \textbf{Enc. GFLOPs} &
\textbf{Latency} & \textbf{CIDEr-D} \\
& \textbf{(M)} & & \textbf{(ms)} & \\
\midrule
HIMEC ECSD-Zero (R50) &
\multicolumn{1}{c}{\textbf{42.8}} &
\multicolumn{1}{c}{\underline{17.9}} & 298 & 136.07 \\
HIMEC DCR-SCST (SwT) &
\multicolumn{1}{c}{\underline{46.2}} &
\multicolumn{1}{c}{\textbf{13.0}} & 333 &
\multicolumn{1}{c}{\textbf{142.81}} \\
RSICCformer            & 201.5 & 46.2 & 32  & 134.19 \\
Chg2Cap                & 329.0 & 70.1 & 24  &
\multicolumn{1}{c}{\underline{136.63}} \\
\bottomrule
\end{tabular}
\par\vspace{1pt}
{\footnotesize Batch size 1 and beam width 3; latency is averaged over 100 passes. R50: ResNet-50; SwT: Swin-T. RSICCformer and Chg2Cap are rescored under our harness. HIMEC uses uncached, unbatched beam search; the baselines use cached, batched decoding. Bold and underline mark first and second for parameters and encoder GFLOPs (lower) and CIDEr-D (higher); latency is unranked because implementations differ.\par}
\end{table}

\subsection{Qualitative Results}

\begin{figure*}[t]
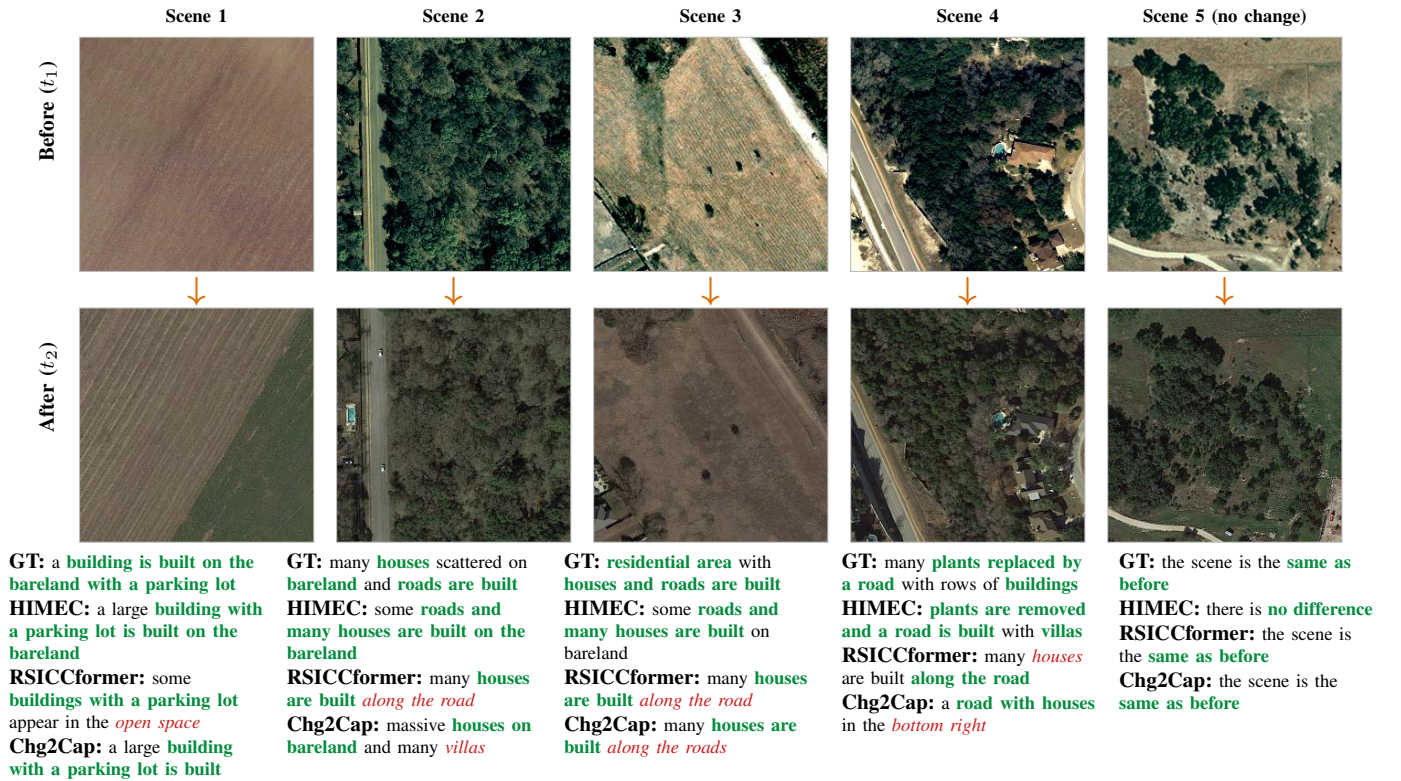

\centering
\scriptsize
\setlength{\fboxsep}{0pt}\setlength{\fboxrule}{0.4pt}
\newcommand{\limg}[1]{\fcolorbox{gray!55}{white}{\includegraphics[width=\dimexpr\linewidth-1pt\relax,height=\dimexpr\linewidth-1pt\relax]{#1}}}
\newcommand{\cml}[2]{{\footnotesize\textbf{#1:}}~#2\par\vspace{1pt}}
\newcommand{\barrow}{\textcolor{orange!85!black}{\large$\boldsymbol{\downarrow}$}}
\setlength{\tabcolsep}{4pt}\renewcommand{\arraystretch}{1.0}
\begin{tabular}{@{}c*{5}{>{\centering\arraybackslash}m{0.172\textwidth}}@{}}
  & \textbf{Scene 1} & \textbf{Scene 2} & \textbf{Scene 3}
  & \textbf{Scene 4} & \textbf{Scene 5 (no change)} \\[3pt]
  \rotatebox{90}{\footnotesize\textbf{Before ($t_1$)}}
    & \limg{compare/b1433_A.png} & \limg{compare/b1713_A.png} & \limg{compare/b744_A.png}
    & \limg{compare/b530_A.png} & \limg{compare/b5_A.png} \\[2pt]
  & \barrow & \barrow & \barrow & \barrow & \barrow \\[2pt]
  \rotatebox{90}{\footnotesize\textbf{After ($t_2$)}}
    & \limg{compare/b1433_B.png} & \limg{compare/b1713_B.png} & \limg{compare/b744_B.png}
    & \limg{compare/b530_B.png} & \limg{compare/b5_B.png} \\
\end{tabular}

\vspace{2pt}
\setlength{\tabcolsep}{4.5pt}
\newcommand{\capcol}[4]{%
  \begin{minipage}[t]{\linewidth}\raggedright\setlength{\parindent}{0pt}\linespread{1.04}\selectfont
    \cml{GT}{#1}\cml{HIMEC}{#2}\cml{RSICCformer}{#3}\cml{Chg2Cap}{#4}
  \end{minipage}}
\begin{tabular}{@{}*{5}{p{0.185\textwidth}}@{}}
  \capcol
    {a \cok{building is built on the bareland with a parking lot}}
    {a large \cok{building with a parking lot is built on the bareland}}
    {some \cok{buildings with a parking lot} appear in the \cerr{open space}}
    {a large \cok{building with a parking lot is built}} &
  \capcol
    {many \cok{houses} scattered on \cok{bareland} and \cok{roads are built}}
    {some \cok{roads and many houses are built on the bareland}}
    {many \cok{houses are built} \cerr{along the road}}
    {massive \cok{houses on bareland} and many \cerr{villas}} &
  \capcol
    {\cok{residential area} with \cok{houses and roads are built}}
    {some \cok{roads and many houses are built} on bareland}
    {many \cok{houses are built} \cerr{along the road}}
    {many \cok{houses are built} \cerr{along the roads}} &
  \capcol
    {many \cok{plants replaced by a road} with rows of \cok{buildings}}
    {\cok{plants are removed and a road is built} with \cok{villas}}
    {many \cerr{houses} are built \cok{along the road}}
    {a \cok{road with houses} in the \cerr{bottom right}} &
  \capcol
    {the scene is the \cok{same as before}}
    {there is \cok{no difference}}
    {the scene is the \cok{same as before}}
    {the scene is the \cok{same as before}} \\
\end{tabular}
\caption{Selected LEVIR-CC captions from HIMEC, RSICCformer, and Chg2Cap. Phrases judged visually supported are shown in \cok{bold green}; errors or unsupported details are shown in \cerr{italic red}. Ground-truth captions are reproduced verbatim from the dataset.}
\label{fig:qual_compare}
\end{figure*}

Fig.~\ref{fig:qual_compare} shows five LEVIR-CC examples. HIMEC describes the parking-lot structure in Scene 1, the houses and roads in Scenes 2 and 3, and the road with nearby buildings in Scene 4. All three methods handle the unchanged pair in Scene 5. The errors shown for the baselines include omitted changes and unsupported locations or objects.

\begin{figure*}[!t]
\centering
\begin{tikzpicture}
\node[inner sep=0,anchor=south west] (dcrvis) at (0,0)
  {\includegraphics[width=0.93\linewidth,keepaspectratio]{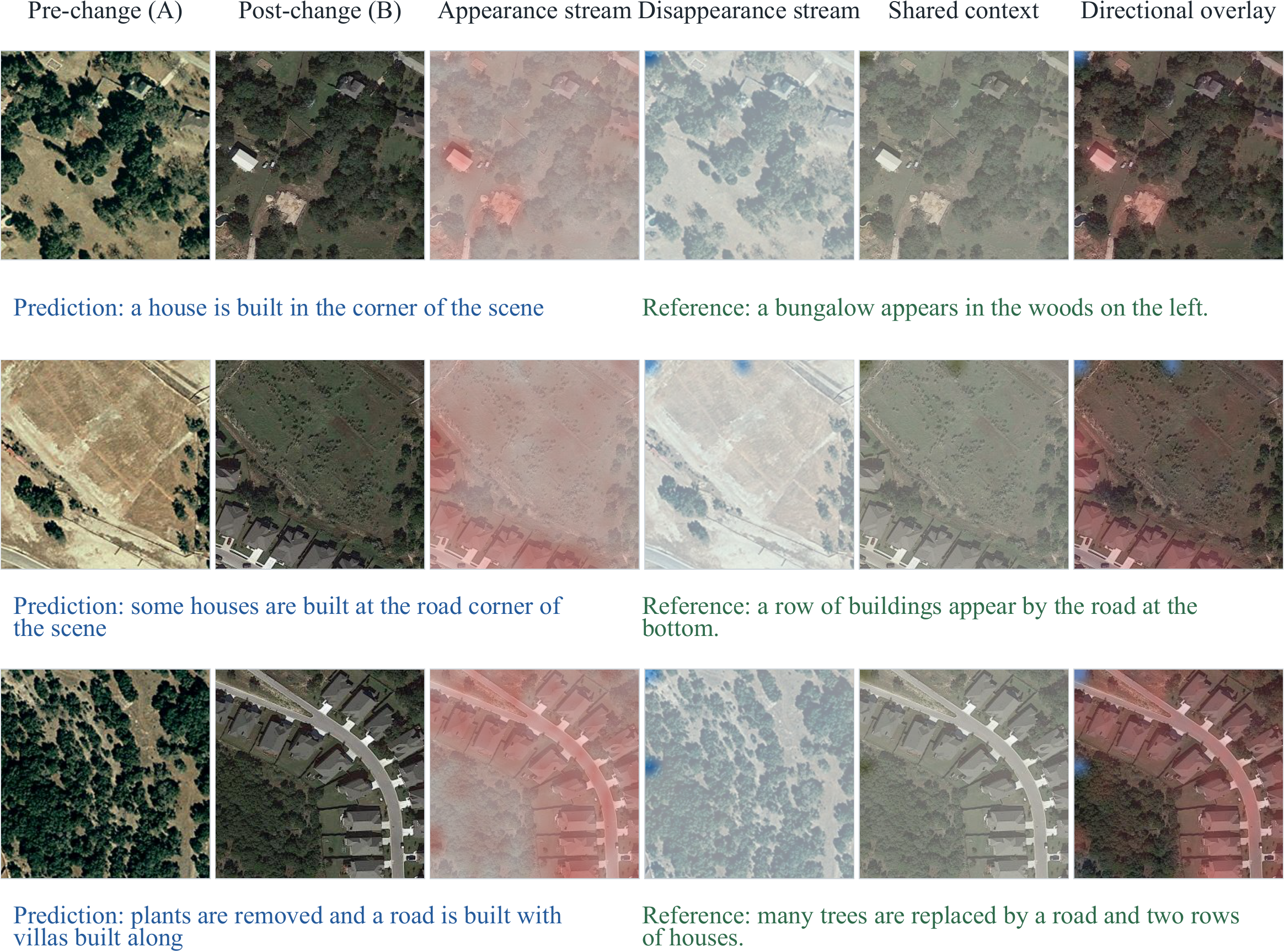}};
\begin{scope}[x={(dcrvis.south east)},y={(dcrvis.north west)}]
  \fill[white] (0,0.947) rectangle (1,1);
  \node[font=\bfseries\footnotesize] at (0.0833,0.975) {Pre-change image};
  \node[font=\bfseries\footnotesize] at (0.2500,0.975) {Post-change image};
  \node[font=\bfseries\footnotesize] at (0.4167,0.975) {Appearance};
  \node[font=\bfseries\footnotesize] at (0.5833,0.975) {Disappearance};
  \node[font=\bfseries\footnotesize] at (0.7500,0.975) {Shared context};
  \node[font=\bfseries\footnotesize] at (0.9167,0.975) {Directional overlay};
\end{scope}
\end{tikzpicture}
\par\vspace{1pt}

\begin{minipage}[b]{0.48\linewidth}
\centering
\begin{tikzpicture}[font=\footnotesize]
\def\CW{2.65}
\def\CH{2.05}
\filldraw[fill=himecGreen!14, draw=black!35, line width=0.8pt]
  (0,\CH) rectangle (\CW,{2*\CH});
\filldraw[fill=himecRed!10, draw=black!35, line width=0.8pt]
  (\CW,\CH) rectangle ({2*\CW},{2*\CH});
\filldraw[fill=himecRed!10, draw=black!35, line width=0.8pt]
  (0,0) rectangle (\CW,\CH);
\filldraw[fill=himecGreen!22, draw=black!35, line width=0.8pt]
  (\CW,0) rectangle ({2*\CW},\CH);
\draw[black!45, line width=0.9pt] (0,0) rectangle ({2*\CW},{2*\CH});
\draw[black!35, line width=0.7pt] (\CW,0) -- (\CW,{2*\CH});
\draw[black!35, line width=0.7pt] (0,\CH) -- ({2*\CW},\CH);

\node[font=\normalsize, text=himecGreen!65!black]
  at ({0.5*\CW},{1.5*\CH+0.16}) {950};
\node[font=\tiny, text=black!65]
  at ({0.5*\CW},{1.5*\CH-0.20}) {True Negative};
\node[font=\normalsize, text=himecRed!85!black]
  at ({1.5*\CW},{1.5*\CH+0.16}) {15};
\node[font=\tiny, text=black!65]
  at ({1.5*\CW},{1.5*\CH-0.20}) {False Positive};
\node[font=\normalsize, text=himecRed!85!black]
  at ({0.5*\CW},{0.5*\CH+0.16}) {85};
\node[font=\tiny, text=black!65]
  at ({0.5*\CW},{0.5*\CH-0.20}) {False Negative};
\node[font=\normalsize, text=himecGreen!65!black]
  at ({1.5*\CW},{0.5*\CH+0.16}) {879};
\node[font=\tiny, text=black!65]
  at ({1.5*\CW},{0.5*\CH-0.20}) {True Positive};

\node[font=\tiny, align=center]
  at ({0.5*\CW},{2*\CH+0.35}) {Predicted:\\No Change};
\node[font=\tiny, align=center]
  at ({1.5*\CW},{2*\CH+0.35}) {Predicted:\\Change};
\node[font=\tiny, rotate=90, align=center]
  at (-0.65,{1.5*\CH}) {Ground Truth:\\No Change};
\node[font=\tiny, rotate=90, align=center]
  at (-0.65,{0.5*\CH}) {Ground Truth:\\Change};
\node[font=\tiny, align=center]
  at ({1.0*\CW},-0.42)
  {F1\,=\,0.9462\quad Precision\,=\,0.9832\quad Recall\,=\,0.9118};
\end{tikzpicture}
\par\centering\footnotesize (a)
\end{minipage}
\hfill
\begin{minipage}[b]{0.48\linewidth}
\centering
\begin{tikzpicture}
\begin{axis}[
  himec,
  width=0.92\linewidth,
  height=5.8cm,
  ybar=2pt,
  bar width=16pt,
  ymin=0,
  ymax=16,
  ytick={0,3,6,9,12,15},
  ylabel={Mean caption length (words)},
  ylabel style={font=\scriptsize},
  symbolic x coords={Changed,Unchanged},
  xtick=data,
  xticklabel style={font=\small},
  extra y ticks={10.89},
  extra y tick labels={Ref.\ 10.89},
  extra y tick style={
    grid=major,
    grid style={dashed,gray!55,thick},
    tick label style={font=\scriptsize,text=gray!65,xshift=2pt}
  },
  nodes near coords,
  nodes near coords style={font=\scriptsize},
  enlarge x limits=0.45,
  legend style={
    at={(0.97,0.97)},
    anchor=north east,
    font=\scriptsize,
    inner sep=1.5pt
  }
]
\addplot[fill=himecBlue!30, draw=himecBlue!75!black]
  coordinates {(Changed,11.49) (Unchanged,4.23)};
\addlegendentry{CE training}
\addplot[fill=himecGreen!35, draw=himecGreen!75!black]
  coordinates {(Changed,9.54) (Unchanged,4.09)};
\addlegendentry{SCST fine-tuning}
\end{axis}
\end{tikzpicture}
\par\centering\footnotesize (b)
\end{minipage}

\caption{DCR visualizations and test-set analyses on LEVIR-CC. The upper rows show the image pair, three DCR responses, their overlay, and captions from the final seed-42 HIMEC checkpoint; references are dataset captions. The bottom panels show (a) its change/no-change confusion matrix over 1,929 pairs and (b) mean word counts for matched seed-42 DCR checkpoints after CE and SCST; the dashed line is the reference mean.}
\label{fig:analysis}
\end{figure*}

Fig.~\ref{fig:analysis} shows strong appearance-oriented responses on new construction and weaker disappearance-oriented responses in the selected examples. The selected failures involve subtle or small-area changes with weak difference signals. For the final seed-42 HIMEC checkpoint, the change head produces 85 false negatives and 15 false positives. Relative to its matched CE checkpoint, SCST reduces the mean changed-caption length from 11.49 to 9.54 words; the dataset-level mean length of changed references is 10.89. These visualizations are used for illustration and do not provide separate causal evidence.

\subsection{Limitations}
\label{sec:limitations}

The matched-regime recovery appears on LEVIR-CC and SECOND-CC but not on DUBAI-CC. The matched 300-pair LEVIR-CC control does not reproduce the DUBAI-CC failure, so sample count alone does not account for the difference within these settings; one subset and one seed cannot identify the relevant dataset or optimization factor. Although the discrepancy recurs across backbones, local decoders, and an independent implementation, the study remains limited to local-to-scene cascades and does not test a published third-party dual-decoder system.

The three-seed comparison isolates query memory from direct fused-feature memory at the complete-system level, and the CE ablations isolate individual components under one seed. They do not isolate DCR as a whole, and the single-seed ablations do not quantify training variance. In particular, the evaluated removals provide no positive evidence for the diversity objective, phrase supervision, two-layer query encoder, or disappearance stream. The complete DCR-SCST system has not been evaluated on a second dataset, and the fixed random type adapter in~\eqref{eq:type_adapter} is retained only for checkpoint reproduction and has not been ablated.

Bootstrap tests are conditional on the trained checkpoints, and cross-method comparisons use nonidentical evaluation pipelines. Automatic metrics do not replace human evaluation of caption factuality and fluency. The parser-derived phrase subsets are post hoc and show substantial disagreement among references. They characterize model behavior under the implemented parser but do not provide human-audited evidence of improved coverage on multi-change scenes.
\section{Conclusion}
\label{sec:conclusion}

HIMEC is a structured change-query framework for remote sensing image change captioning. DCR organizes signed bitemporal evidence into appearance-oriented, disappearance-oriented, and shared-context streams before learned fusion. The resulting query tokens form the only sample-dependent semantic memory supplied to the adopted scene decoder. A caption-derived phrase objective is used only as auxiliary supervision during cross-entropy training; the local phrase decoder is omitted from self-critical training and inference.

The local-to-scene cascade was evaluated separately as a diagnostic baseline. Regime-matched conditioning recovered most of the observed loss on LEVIR-CC and SECOND-CC; content permutation, which removes image-state correspondence, produced no detectable penalty. These controls are consistent with a contribution from conditioning inconsistency in the evaluated cascades, but they do not establish a field-wide RSICC limitation or equivalence among the interventions. The fixed zero input disables rather than aligns the local-state pathway.

With SCST, HIMEC achieves $142.81\!\pm\!0.60$ CIDEr on LEVIR-CC, compared with $139.51\!\pm\!3.40$ for matched direct fused-feature memory. In the configuration-matched single-seed CE ablations, scores decrease when the difference interaction, appearance stream, or shared context is removed, whereas removing several other retained components does not reduce the observed score. Cross-method differences remain descriptive because published results use different evaluation pipelines. Remaining evidence gaps include multi-seed component ablations, removal of DCR as a complete block, evaluation of the complete model on another dataset, human-audited phrase counts, third-party cascaded architectures, and human assessment of caption factuality and coverage.

\newpage

\appendices
\section{Legacy-Scorer Diagnostic Tables}
The following tables are retained for within-harness diagnostic comparisons only. They are separated from the standard PTBTokenizer results in the main text to prevent cross-table metric comparisons.

\begin{table}[!t]
\centering
\caption{Diagnostic comparison of local-state conditioning interventions on the LEVIR-CC test set using ResNet-50, seed 42, and the legacy scorer. TF/AR denotes teacher-forced training and autoregressive inference; SS and PF denote scheduled sampling and professor forcing, respectively.}
\label{tab:mismatch}
\tstrut
\setlength{\tabcolsep}{5pt}
\footnotesize
\begin{tabular}{@{}lS[table-format=3.2]S[table-format=+2.2]@{}}
\toprule
\textbf{Variant} & {\textbf{CIDEr}} & {$\boldsymbol{\Delta}$ \textbf{vs. baseline}} \\
\midrule
Cascaded TF/AR baseline              & 121.79 & {\text{N/A}} \\
SS annealed (linear)                 & 115.46 & {-6.33}  \\
Professor Forcing                    & 120.29 & {-1.50}  \\
SS ($p{=}0.25$)                      & 128.00 & {+6.21}  \\
SS ($p{=}0.5$)                       & 128.91 & {+7.12}  \\
SS ($p{=}0.75$)                      & 131.03 & {+9.24}  \\
Greedy-Aligned                       & 133.36 & {+11.57} \\
Matched-Path                         & 134.44 & {+12.65} \\
Content-Scrambled &
\multicolumn{1}{c}{\underline{135.36}} &
\multicolumn{1}{c}{\underline{+13.57}} \\
ECSD-Zero$^{\star}$ &
\multicolumn{1}{c}{\textbf{136.59}} &
\multicolumn{1}{c}{\textbf{+14.80}} \\
\bottomrule
\end{tabular}
\par\vspace{1pt}
{\footnotesize All rows use the same legacy scorer without PTBTokenizer, so only within-table differences are compared. $^{\star}$Adopted interface. Bold and underline mark the top two values within this diagnostic.\par}
\end{table}

\begin{table}[!t]
\centering
\caption{Sensitivity to image perturbations on the LEVIR-CC test set using ResNet-50 and the legacy scorer.}
\label{tab:robust}
\tstrut
\setlength{\tabcolsep}{3pt}
\footnotesize
\begin{tabular}{@{}llS[table-format=3.2]S[table-format=1.4]
                   S[table-format=+2.1]S[table-format=3.2]S[table-format=1.4]@{}}
\toprule
\multirow{2}{*}{\textbf{Perturbation}} &
\multirow{2}{*}{\textbf{Sev.}} &
\multicolumn{3}{c}{\textbf{ECSD-Zero}} &
\multicolumn{2}{c}{\textbf{ECSD-Aug}} \\
\cmidrule(lr){3-5}\cmidrule(lr){6-7}
& & {\textbf{CIDEr}} & {\textbf{CF1}} & {$\boldsymbol{\Delta}$ \textbf{CIDEr}}
  & {\textbf{CIDEr}} & {\textbf{CF1}} \\
\midrule
Clean             & N/A & 135.18 & 0.9374 & {\text{Ref.}} & 136.29 & 0.9389 \\
Brightness (low)  & 0.5 & 135.28 & 0.9360 & {+0.1} & 135.74 & 0.9370 \\
Brightness (high) & 1.0 & 135.25 & 0.9328 & {+0.1} & 135.85 & 0.9370 \\
Contrast (high)   & 1.0 & 130.73 & 0.9353 & {-4.5} & 136.16 & 0.9392 \\
Blur (low)        & 1.0 & 131.44 & 0.7913 & {-3.7} & 136.20 & 0.9378 \\
Blur (medium)     & 2.0 & 116.19 & 0.3960 & {-19.0} & 134.98 & 0.8978 \\
Blur (high)       & 3.0 & 111.29 & 0.2457 & {-23.9} & 130.72 & 0.8012 \\
Translation (sm.) & 1.0 & 135.71 & 0.9410 & {+0.5} & 136.56 & 0.9380 \\
\bottomrule
\end{tabular}
\par\vspace{1pt}
{\footnotesize Only within-table changes are comparable. ECSD-Aug uses independent Gaussian blur for the training pair and shared brightness/contrast jitter. Each test perturbation is shared by both images.\par}
\end{table}

\FloatBarrier

\raggedbottom
\newcommand{\authorphoto}[1]{%
  \parbox[c][1.25in][t]{1in}{\centering%
    \vspace*{-2.5pt}
    \includegraphics[width=1in,height=1.25in,clip,keepaspectratio]{#1}}}
\vspace{-16pt}
\begin{IEEEbiography}[{\authorphoto{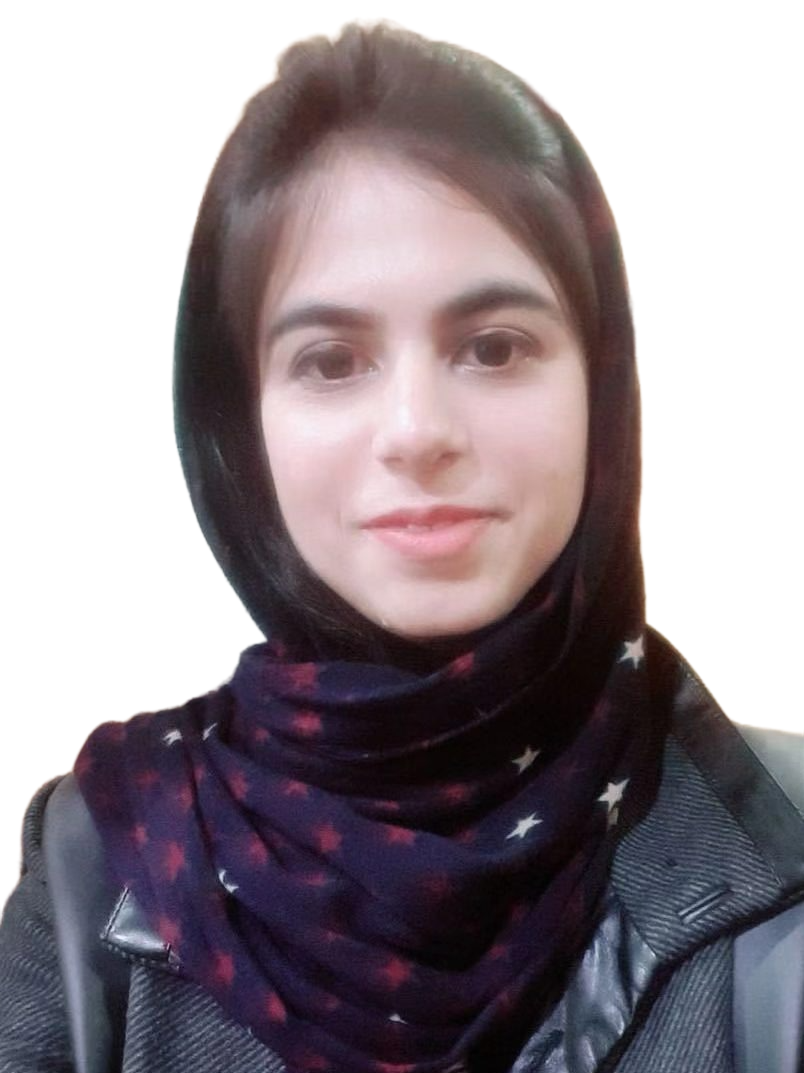}}]{Aysha Ashraf}
received the M.S. degree in computer science and is currently pursuing the Ph.D. degree with the University of Electronic Science and Technology of China (UESTC), Chengdu, China.

Her research interests include remote sensing, computer vision, and change detection in satellite imagery.
\end{IEEEbiography}

\vspace{-16pt}
\begin{IEEEbiography}[{\authorphoto{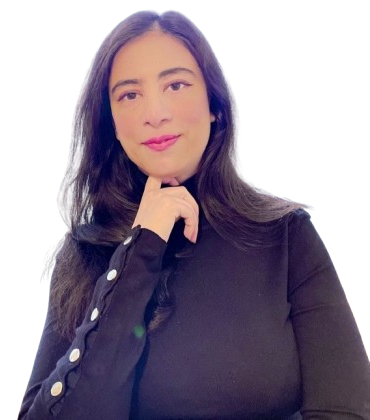}}]{Shaina Ashraf}
received the M.S. degree and is currently pursuing the Ph.D. degree with the Bonn-Aachen International Center for Information Technology (B-IT), University of Bonn, Bonn, Germany.

Her research interests include large language models, machine learning, and natural language processing.
\end{IEEEbiography}

\vspace{-16pt}
\begin{IEEEbiography}[{\authorphoto{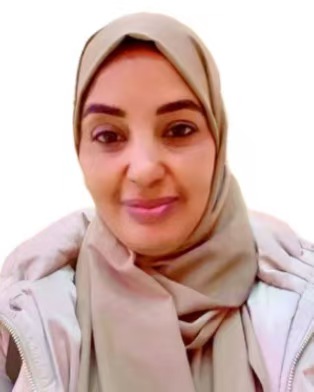}}]{Wafaa I. M. Hussin}
received the M.S. degree in computer architecture and networking from the Faculty of Engineering, University of Khartoum, Sudan, in 2019. She is currently pursuing the Ph.D. degree in information and communication engineering with UESTC, Chengdu, China.

Her research interests include image processing, computer vision, and change detection.
\end{IEEEbiography}

\vspace{-16pt}
\begin{IEEEbiography}[{\authorphoto{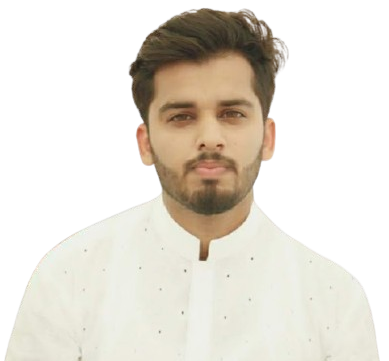}}]{Ali Haider}
received the M.S. degree in information and communication engineering and plans to pursue the Ph.D. degree in information and communication engineering with UESTC, Chengdu, China.

His research interests include remote sensing image analysis, deep learning, and environmental change detection.
\end{IEEEbiography}

\vspace{-16pt}
\begin{IEEEbiography}[{\authorphoto{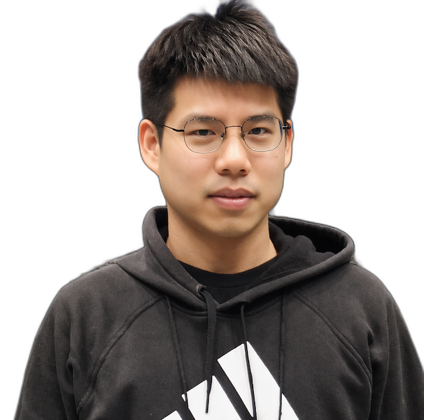}}]{Zhi Lu}
received the B.S., M.S., and Ph.D. degrees from the University of Electronic Science and Technology of China, Chengdu, China, in 2015, 2018, and 2022, respectively. He was a Postdoctoral Researcher and then an Associate Research Fellow at the University of Science and Technology of China from 2023 to 2026. He is currently an Associate Professor with the Laboratory of Intelligent Collaborative Computing, University of Electronic Science and Technology of China.

His research interests include computer vision, multimedia content understanding, and remote sensing image analysis.
\end{IEEEbiography}

\vspace{-24pt}
\begin{IEEEbiography}[{\authorphoto{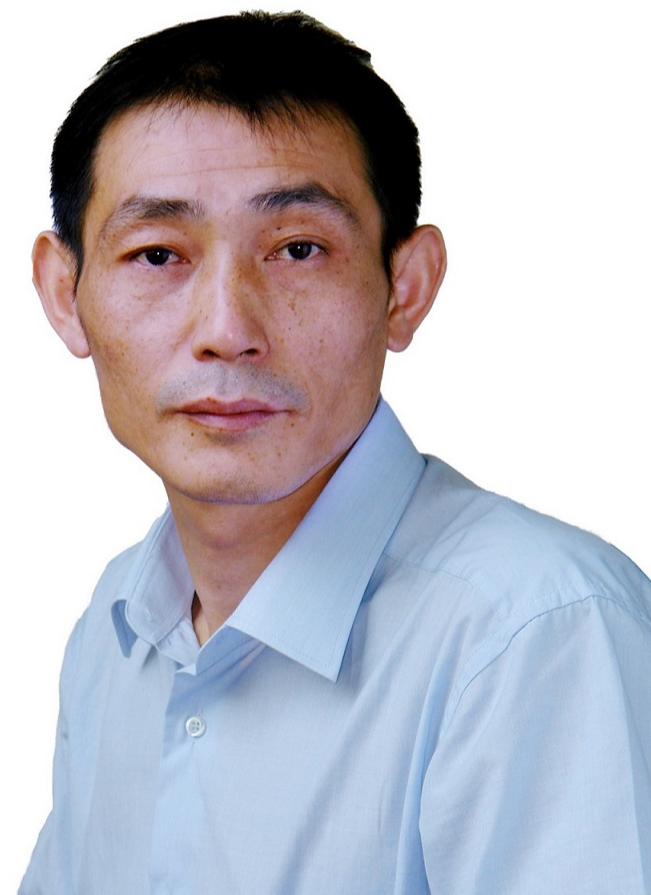}}]{Zhenming Peng}
(Senior Member, IEEE) received the Ph.D. degree in geodetection and information technology from Chengdu University of Technology, China, in 2001.

He is currently a Professor with UESTC, Chengdu. His research interests include image and signal processing and target recognition in remote sensing.
\end{IEEEbiography}

\vfill
\end{document}